N° D'ordre : ……………
Série : …………………

# Mémoire

Pour l'obtention du diplôme de

## MASTER EN INFORMATIQUE

Option: *Architectures Distribuées*

# Thème

## Une Nouvelle Approche de Planification Distribuée Dynamique : Application aux problèmes DPDP

( Dynamic Pick Up and Delivery Problems)

*Présenté par* : **TOLBA Zakaria**

*Devant le jury composé de :*

| | | | |
|---|---|---|---|
| **Président :** | **Mokhati Farid** | Prof | Univ. Oum El Bouaghi |
| **Examinateurs:** | **Marir Toufik** | MCB | Univ. Oum El Bouaghi |
| | **Kouah Sofia** | MCB | Univ. Oum El Bouaghi |
| **Encadreur :** | **Guerram Tahar** | MCB | Univ. Oum El Bouaghi |

2016/2017

بِسْمِ اللهِ الرَّحْمٰنِ الرَّحِيْمِ

# Remerciements

*Mes remerciements, les plus vifs et les plus chaleureux, ainsi que toute ma reconnaissance, sont dédiés à mon encadreur Dr **Tahar GUERRAM** qui a dirigé et supervisé ce travail avec beaucoup d'expérience.*

*Mes sincères remerciements au Mr le Professeur **Farid MOKHATI**, Prof à l'université d'Oum El Bouaghi, qui a accepté de présider le jury*

*Mes remerciements vont aussi aux membres du jury composé de :*

   *Dr. **Toufik MARIR**, MCB à l'université d'Oum El Bouaghi.*

   *Dr **Sofia KOUAH**, MCB à l'université d'Oum El Bouaghi.*

*Tout en les remerciant d'avoir accepté de juger notre travail, nous leur promettons d'accorder une importance particulière à leurs remarques constructives et leurs suggestions pour l'amélioration et l'enrichissement de ce travail.*

*Je remercie également tous les membres du laboratoire ReLa(CS)² pour leur soutien inconditionnel.*

*Je remercie tous ceux qui ont contribué de près ou de loin à l'élaboration de ce travail.*

# Dédicace

*Je dédie ce travail à tous les membres de ma famille*

# ABSTRACT

In this work, we proposed a new dynamic distributed planning approach that is able to take into account the changes that the agent introduces on his set of actions to be planned in order to take into account the changes that occur in his environment. Our approach fits into the context of distributed planning for distributed plans where each agent can produce its own plans. According to our approach the generation of the plans is based on the satisfaction of the constraints by the use of the genetic algorithms.

Our approach is to generate, a new plan by each agent, whenever there is a change in its set of actions to plan. This in order to take into account the new actions introduced in its new plan. In this new plan, the agent takes, each time, as a new action set to plan all the old un-executed actions of the old plan and the new actions engendered by the changes and as a new initial state; the state in which the set of actions of the agent undergoes a change. In our work, we used a concrete case to illustrate and demonstrate the utility of our approach.

# RÉSUMÉ

Dans ce travail, nous avons proposé une nouvelle approche de planification distribuée dynamique, capable de prendre en considération les changements que l'agent introduit sur son ensemble des actions à planifier dans le but de prendre en compte les changements qui surviennent dans son environnement. Notre approche s'intègre dans le contexte de la planification distribuée pour des plans distribués où chaque agent peut produire ses propres plans. Selon notre approche la génération des plans est basée sur la satisfaction des contraintes par l'utilisation des algorithmes génétiques.

Notre approche consiste à faire générer, un nouveau plan par chaque agent, à chaque fois qu'il y'a un changement dans son ensemble d'actions à planifier. Ceci dans le but de prendre en considération les nouvelles actions introduites dans son nouveau plan. Dans ce nouveau plan l'agent prend, à chaque fois, comme nouvel ensemble d'actions à planifier l'ensemble des anciennes actions non exécutées de l'ancien plan et les nouvelles actions engendrées par les changements et comme nouvel état initial; l'état dans lequel l'ensemble des actions de l'agent subit un changement. Dans le cadre de notre travail, nous avons utilisé un cas concret pour illustrer et montrer l'utilité de notre approche.

**Mots-clés**

Agent, Système Multi-agents, Algorithme Génétique, Planification.

# ملخص

في هذا العمل، اقترحنا نهجا جديدا لتخطيط ديناميكي موزع، قادر على الأخذ بعين الاعتبار التغييرات التي يدخلها الوكيل على الإجراءات التي يريد التخطيط لها و هذا من اجل الأخذ بعين الاعتبار التغيرات التي قد تحدث في بيئته. نهجنا يدمج في سياق التخطيط موزع من اجل مخططات موزعة حيث فيها كل وكيل يمكن له وضع مخططه الخاصة يعتمد نهجنا في إنشاء المخططات على مراعاة بعض الشروط باستخدام الخوارزميات الوراثية

نهجنا يحث على إيجاد مخطط جديدة من قبل كل وكيل، في كل مرة كان هناك تغيير في مجموعته من الإجراءات. هذا من أجل الأخذ بعين الاعتبار الإجراءات الجديدة التي أدخلت في الخطة الجديدة. في هذا المخطط الجديد، يأخذ الوكيل، في كل مرة، كمجموعة جديدة من الإجراءات الإجراءات التي لم تتحقق في المخطط القديم و الجديدة الناتجة عن التغير و كحالة جديدة له و لكل النظام الحالة التي تم فيها التغيير

في هذا العمل استعملنا مثالا ملموسا لتوضيح وبيان فائدة نهجنا

**الكلمات المفتاحية**

وكيل ، نظام وكلاء متعدد، الخوارزميات الوراثية، التخطيط

# Table des matières

# Table des figures

# Introduction générale

La planification est une problématique centrale de l'Intelligence Artificielle dont l'objectif consiste à générer un plan d'actions à un niveau symbolique à partir d'un état initial pour atteindre un but défini auparavant [1]. Cependant, en Intelligence Artificielle, la planification pose de multiples problèmes, notamment, ceux liés à la formalisation et au raisonnement qui porte sur l'action, le plan, le changement, le temps et les objectifs à atteindre. Elle pose aussi des problèmes liés à la robustesse pour la prise en compte d'état du monde partiellement connu ou des actions non déterministes, des problèmes algorithmiques pour la génération des plans ainsi que des problèmes de contrôle d'exécution, de réactivité ou d'évolution imprévue de l'environnement et d'adaptation des plans déjà produits.

La planification, dans sa version classique a connu un essor considérable à cause de la richesse des langages de modélisation et l'efficacité des systèmes de génération des plans. Néanmoins, la planification classique souffrait d'une faiblesse causée par le fait qu'elle reposait sur deux hypothèses simplificatrices fortes à savoir : la disposition d'une connaissance parfaite, à tout instant, de l'état du système et des effets des actions et la certitude que les modifications de l'état du système proviennent uniquement de l'exécution des actions du plan. Pour pallier à cette faiblesse, le domaine de la planification dans l'incertain [2] s'est développé, proposant d'intégrer des actions à effet probabiliste puis des fonctions d'utilité additives sur les buts, conduisant à une famille d'approches pour la planification basées sur la théorie de la décision [3] et utilisant des langages de représentation traditionnellement connus en intelligence artificielle: logique, contraintes ou réseaux bayésiens. L'utilisation de ces langages de représentation a fait exploser une grande complexité dans les algorithmes de génération des plans dont la résolution est devenue un défi pour la communauté de l'intelligence artificielle. C'est ainsi que l'idée de l'utilisation des systèmes multi agents a germé chez les chercheurs.

En effet, l'extension de la planification dans le cadre des systèmes multi-agent a aboutit à la planification distribuée [4], [11] dans laquelle le domaine de planification est réparti sur un ensemble d'agents. Ces agents peuvent être coopératifs dans le sens où ils ont un objectif global commun et des capacités complémentaires pour le réaliser ou individualistes dans le sens où ils ont des objectifs individuels dont ils sont capables d'assurer la réalisation sans aide externe. Dans les deux cas les agents doivent être capables de générer des plans qui permettent la réalisation soit des sous-objectifs nécessaires pour un objectif global soit des objectifs individuels. Dans la littérature, il existe quelques travaux sur planification distribuée. Nous citons, entre autres [177], [178], [179], [180].

Le paradigme agent revêt de plus en plus d'importance pour sa capacité à aborder les systèmes complexes caractérisés par l'indéterminisme, l'émergence et l'évolution imprédictible. Il est très efficace pour gérer la nature hétérogène des composantes d'un système, pour modéliser les interactions entre les composantes de ce dernier et pour tenter de comprendre les phénomènes émergents qui en découlent. Ceci est lié au fait que l'agent possède un comportement, caractérisé principalement par quatre propriétés [5]:

- Autonomie ou proactivité : capacité à agir sans intervention extérieure, prise d'initiative.
- Sensibilité : capacité à percevoir l'environnement ou les autres agents.
- Localité : limitation de la perception et des actions.
- Flexibilité : réaction aux changements perçus.

En effet, l'agent ne se limite pas seulement à réagir aux invocations de méthodes spécifiques, comme il est souvent le cas dans le paradigme objet, mais également à tout autre changement observable dans son environnement. La prise en compte de ces changements se traduit automatiquement par un ensemble d'actions nouvelles que l'agent doit exécuter. La détermination de ces actions dépend de la nature de l'agent [6]. En effet si, par exemple, l'agent est rationnel, les actions à déterminer ne doivent pas être en opposition avec la fonction d'utilité de l'agent, si l'agent est avec but, ces actions ne doivent pas être en opposition avec le but de l'agent, si l'agent est réactif avec modèle, ces actions sont prédéterminée par un ensemble de règles, etc.

Le comportement de l'agent est ainsi source d'avantages mais les actions nouvelles à exécuter par l'agent, afin de prendre en considération les changements imprédictibles qui caractérisent son environnement, peuvent créer un problème lors de la planification distribuée. En effet dans la planification classique, l'ensemble des actions à planifier est défini auparavant et ne subit aucun changement assurant ainsi, une fiabilité du plan généré jusqu'à la fin de son exécution. Par contre dans la planification distribuée, chaque agent peut avoir des changements dans son ensemble d'actions à planifier, suite aux changements imprédictibles de son environnement. En effet à cause des changements survenus sur l'ensemble des actions, le plan que l'agent était entrain d'exécuter devient obsolète car il ne prend pas en considération les nouvelles actions à exécuter par l'agent, afin de prendre en considération les changements imprédictibles de son environnement. L'agent se trouve par conséquent, contraint de générer un nouveau plan. De ce fait, la réflexion vers une approche de planification dynamique permet de générer, à tout moment et au fur et à mesure des changements, de nouveaux plans pour prendre en considération les nouvelles actions s'impose d'elle-même. Ceci représente le point focal de notre travail, dans lequel, nous proposons une nouvelle approche de planification dynamique distribuée capable de prendre en considération les changements pouvant survenir sur l'ensemble des actions à planifier. Notre approche s'intègre dans le contexte de la planification distribuée pour des plans distribués où chaque agent peut produire ses propres plans. Selon notre approche la génération des plans est basée sur la satisfaction des contraintes par l'utilisation des algorithmes génétiques.

Le présent mémoire est organisé en quatre chapitres structurés comme suit :

- **Système multi-agents :** une synthèse de l'état de l'art sur les systèmes multi-agents est décrite dans ce chapitre qui se compose de deux parties :
    - La première partie concerne l'agent à savoir : sa définition, ses caractéristiques, les types d'agents.
    - L'autre partie concerne les systèmes multi agents à savoir : leur définition, leurs caractéristiques, leur modélisation, leur méthodologie, leur implémentation et leur domaine d'application.
- **Planification multi-agents :** une synthèse de l'état de l'art la planification est décrite dans ce chapitre qui se compose de quatre parties :

    - La première partie concerne la planification classique à savoir : sa définition, ses représentation, ses algorithmes.

- La deuxième partie concerne la planification hiérarchique.
- La troisième partie concerne la planification dans l'incertain à savoir : sa définition, ses représentations, ses algorithmes.
- La quatrième partie concerne la planification multi-agents à savoir : sa définition, les types de planification multi-agents, la coordination des plans.

- **L'approche proposée**: l'approche de planification dynamique proposée qui prend en considération les changement imprédictibles dans l'ensemble d'actions à planifier est décrite dans ce chapitre.
- **Étude de cas** : un cas concret sur lequel a été appliquée l'approche et qui montre la faisabilité et les intérêts de cette dernière est traitée dans ce chapitre.

# Chapitre 1
# Les systèmes multi-agents

### 1. Introduction

L'approche orientée-agent a été introduite au début des années 90 [7], [8], elle découle principalement du domaine de l'IAD mais s'inspire aussi de domaines très diversifiés tels que le génie logiciel, les systèmes répartis et les sciences sociales.

La technologie orientée-agent est devenue un paradigme à part entière du génie logiciel disposant de ses propres éléments méthodologiques en termes de conception et de programmation. En effet, ces deux dernières décennies ont été marquées par l'apparition d'un très grand nombre de méthodologies et de plates-formes orientées agent. Ainsi, les systèmes multi-agents revêtent de plus en plus d'importance pour leur capacité à aborder les systèmes complexes caractérisés par l'indéterminisme, l'émergence et l'évolution imprédictible. Ils sont très efficaces pour gérer la nature hétérogène des composantes d'un système, pour modéliser les interactions entre ses composantes et pour tenter de comprendre les phénomènes émergents qui en découlent.

Ce chapitre présente une synthèse de l'état de l'art sur les systèmes multi-agents. Il se compose de deux parties ; la première partie concerne l'agent à savoir ; sa définition, ses caractéristiques, les types d'agents. L'autre partie concerne les systèmes multi agents à savoir : leur définition, leurs caractéristiques, leur modélisation, leur méthodologie, leur implémentation et leur domaine d'application.

## 2. Agent
### 2.1. Définition

Plusieurs travaux ont porté sur la notion d'agent engendrant ainsi des définitions aussi riches que variées. Parmi ces définitions on trouve :

**[Ferber 91]** [9]: L'agent est une entité physique ou virtuelle mue par un ensemble de tendances (objectifs individuels, fonction de satisfaction ou de survie à optimiser), qui possède des ressources propres, ne dispose que d'une représentation partielle (éventuellement aucune) de son environnement, son comportement tendant à satisfaire ses objectifs, en tenant compte de ses ressources et de ses compétences, et en fonction de sa perception, ses représentations et ses communications.

**[Ferber, 95]** [10] : L'agent est une entité physique ou virtuelle :

- qui est capable d'agir dans un environnement ;
- qui peut communiquer directement avec d'autres agents ;
- qui est mue par un ensemble de tendances (sous la forme d'objectifs individuels ou de fonctions de satisfaction, voire de survie, qu'elle cherche à optimiser) ;
- qui possède des ressources propres ;
- qui est capable de percevoir son environnement (mais de manière limitée) ;
- qui ne dispose que d'une représentation partielle de cet environnement (et éventuellement aucune) ;
- qui possède des compétences et offre des services ;
- qui peut éventuellement se reproduire ;
- et dont le comportement tend à satisfaire ses objectifs, en tenant compte des ressources et des compétences dont elle dispose, et en fonction de sa perception, de ses représentations et des communications qu'elle reçoit.

**[P. Maes, 95]** [12]: Un agent autonome est un système calculatoire qui, placé dans un environnement complexe et dynamique, perçoit et agit de manière autonome dans cet environnement et, ce faisant, réalise les objectifs ou des tâches pour lesquels il est conçu.

**[Jennings et Wooldridge, 98]** [13] : Un agent est un système informatique situé dans un certain environnement, capable d'exercer de façon autonome des actions sur cet environnement en vue d'atteindre ses objectifs.

**[DeLoach et al, 01]** [14] : Un agent est un ensemble de processus qui communiquent entre eux pour atteindre un objectif donné.

### 2.2. Caractéristiques d'un agent

Plusieurs caractéristiques sont reliées aux agents [15]. Un agent ne possède pas forcément toutes ces propriétés. Parmi ces caractéristiques on trouve :

- **Autonomie** : un agent possède un état interne (non accessible aux autres) en fonction duquel il entreprend des actions sans intervention d'humains ou d'autres agents.
- **Réactivité** : un agent perçoit des stimuli provenant de son environnement et réagit en fonction de ceux-ci.
- **La pro-activité** : la capacité de l'agent de prendre l'initiative en démontrant des comportements orientés objectifs.
- **Capacité à agir** : un agent est mû par un certain nombre d'objectifs qui guident ses actions, il ne répond pas simplement aux sollicitations de son environnement.
- **Sociabilité** : un agent communique avec d'autres agents ou des humains et peut se trouver engager dans des transactions sociales (négocier ou coopérer pour résoudre un problème) afin de remplir ses objectifs.
- **Flexibilité** : la flexibilité peut être vue comme une forme de l'intelligence. Etre flexible signifie que l'agent est [16] :
  - ✓ **Réactif** : l'agent doit être capable de percevoir son l'environnement et de répondre à temps aux changements qui peuvent affecter cet environnement.
  - ✓ **Proactif** : l'agent n'agit pas seulement d'une manière réactive (en fonction de son environnement) mais il doit avoir un comportement orienté objectif et que l'agent peut prendre des initiatives.
  - ✓ **Social** : l'agent est capable d'interagir avec les autres agents intelligents et humains pour qu'il puisse atteindre ses propres objectifs et aider les autres dans leurs activités.

- **Adaptabilité** : un agent adaptatif est un agent capable de modifier son comportement et ses objectifs en fonction de ses interactions avec son environnement et avec les autres agents [17].
- **Intentionnalité** : un agent intentionnel est un agent guidé par ses buts. Une intention [18] est la déclaration explicite des buts et des moyens d'y parvenir. Elle exprime donc la volonté d'un agent d'atteindre un but ou d'effectuer une action.
- **Rationalité** : un agent rationnel est un agent qui suit le principe suivant [19] : « Si un agent sait qu'une de ses actions lui permet d'atteindre un de ses buts, il la sélectionne ». Les agents rationnels disposent de critères d'évaluation de leurs actions, et sélectionnent selon ces critères les meilleures actions qui leur permettent d'atteindre le but.
- **La perception** : la perception est un moyen de recevoir de l'information d'un environnement qui est souvent complexe et évolutif.
- **La contribution** : l'agent participe plus ou moins à la résolution du problème ou à l'activité globale du système.
- **Le contrôle** : il peut-être totalement distribué entre les agents mais peut être voué à une certaine classe d'agents comme les agents « facilitateurs ».

### 2.3. Types d'agent

Les différentes propriétés des agents sont souvent utilisées comme critères de classification. Plusieurs classifications ont été proposées parmi lesquelles on trouve :

- **[Wooldridge et Jennings, 95]** [20] proposent deux classifications pour les agents :
    - ✓ **Agent faible :** Définit comme une entité physique ou logicielle autonome, réactive, proactive et sociable.
    - ✓ **Agent fort :** Conceptuellement un agent fort possède des propriétés mentales et émotionnelles comme : les connaissances, les croyances, les intentions et les obligations.

- **[Moulin et Chaib-Draa**, **96]** [21] caractérisent les agents par leur capacité à résoudre les problèmes :

    - ✓ **Agent réactif :** Les agents réactifs sont une catégorie spéciale d'agents qui ne disposent pas de modèles internes et symboliques de leur environnement.

Leurs capacités consistent à réagir uniquement aux stimuli provenant de l'environnement. Ils sont relativement simples et interagissent avec les autres agents de façon basique.

- ✓ **Agent intentionnel :** capable de raisonner à propos de ses intentions, ses croyances et peut concevoir et exécuter des plans d'actions.
- ✓ **Agent social** : possède en plus des propriétés d'un agent intentionnel, des modèles explicites des autres agents.

- **[Ferber ,95]** [22] a proposé deux classifications pour les agents :
    - ✓ Selon le caractère :
        - ➢ **Agent cognitif** : possède une représentation (partielle mais sophistiquée) de leur environnement, ont des buts explicites et sont capables de planifier leur comportement, de mémoriser leurs actions passées, de communiquer par envoi de messages ou via des langages d'interaction élaborés, de négocier, etc. Les agents cognitifs peuvent être intentionnels (dotés d'attitudes intentionnelles telles que les croyances, les désirs et les intentions), rationnels (agissant selon une rationalité donnée telle que la rationalité économique), normés (agents évoluant dans un système doté de normes sociales), etc.
        - ➢ **Agent réactif** : (voir plus haut).
    - ✓ Selon le comportement :
        - ➢ **Agent téléonomique** : dirigé vers des buts explicites.
        - ➢ **Agent réflexe** : régi par des perceptions.

- **[Nwana, 96]** [23] propose une typologie des agents à partir de plusieurs critères de classification :
    - ✓ Mobilité :
        - ➢ **Agent statique :** Un agent statique est un agent qui s'exécute seulement sur le système où il est créé. Pour la récupération des informations supplémentaires inexistantes localement il utilise le mécanisme RPC (Remote Procedure Calling) qui est une méthode appliquée dans le modèle client serveur [48].
        - ➢ **Agent mobile :** Un agent mobile est un agent capable de se déplacer dans son environnement, qui peut être physique (réel ou simulé) ou structurel (niveaux d'exécution par exemple).

- ✓ Présence d'un modèle de raisonnement symbolique :
  - ➢ **Agent délibératif :** Un agent délibératif est un agent qui détient une représentation symbolique du monde et choisi les actions à accomplir à l'aide d'un raisonnement logique.
  - ➢ **Agent réactif :** (voir plus haut).
- ✓ Existence d'un objectif et de propriétés initiales comme l'autonomie, la coopération et l'apprentissage. A partir de ces propriétés, quatre types d'agents sont déduits:
  - ➢ **Agent collaboratifs :** Les agents collaboratifs ont davantage des caractéristiques d'autonomie et de coopération avec les autres agents dans la réalisation de leurs objectifs. Ils doivent pouvoir négocier afin d'arriver à des compromis acceptables.
  - ➢ **Agent collaboratifs et apprenants :** c'est un agent collaboratif capable d'apprendre.
  - ➢ **Agent d'interface :** Ces agents collaborent avec l'utilisateur pour effectuer certaines tâches. La collaboration de ces agents avec l'utilisateur ne nécessite pas l'existence explicite d'un langage de communication. Les agents d'interface apprennent les préférences des utilisateurs afin de mieux les assister.
  - ➢ **Agent intelligents** : Un agent intelligent est une entité logicielle qui réalise des opérations pour le compte d'un usager (ou d'un autre programme) avec un certain degré de liberté et d'autonomie et qui, pour ce faire, exploite des connaissances ou des représentations des désirs et des objectifs de l'usager.
- ✓ Rôles : recherche d'informations ou travail sur Internet ….
  - ➢ **Agents d'interface :** (voir plus haut).
  - ➢ **Agents pour la recherche d'informations :** Ces agents effectuent, en premier lieu, une recherche d'informations parmi une collection de données et, en second lieu, procèdent à une analyse des informations utiles trouvées afin de découvrir de nouvelles connaissances.
  - ➢ **Agents pour le commerce électronique :** La montée de l'internet a bien entendu crée de nouvelles nécessitées. Les agents issus de cette tendance permettent la promotion, la vente ainsi que l'achat de produits et de services par l'entremise des réseaux informatiques, etc.

> **Agents conversationnels animés :** Ce sont des interfaces de dialogue entre des utilisateurs et des systèmes d'information. Ils se déploient sur des sites Internet, notamment des sites marchands. Ils sont pourvus de bases de dialogues correspondant aux contextes d'interaction dans lesquels ils agissent.

> **Agents guides ou assistants :** Ce type d'agents essaye de suggérer des sites susceptibles d'intéresser l'usager, en observant son comportement sur Internet. Le principe est assez simple. Généralement les sites que l'on visite sur Internet reflètent les goûts ou les besoins de l'utilisateur. Ainsi, en analysant ses habitudes de navigation, un « agent assistant » tente d'apprendre des habitudes de son utilisateur et de lui suggérer des sites en relation avec ce qu'il désire.

- **[Savall, 03]** [24] classe les agents selon le degré d'intelligence :
    - ✓ **Agents cognitifs :** (voir plus haut).
    - ✓ **Agents réactifs :** (voir plus haut).
    - ✓ **Agents hybrides :** Un agent hybride consiste en la combinaison plusieurs caractéristiques au sein d'un même agent ; ces caractéristiques concernent la mobilité, la collaboration, l'autonomie, la capacité à apprendre, etc.
- **[Franklin et Graesser , 96]** [25] proposent une typologie des agents qui est présentée dans la figure ci-après :

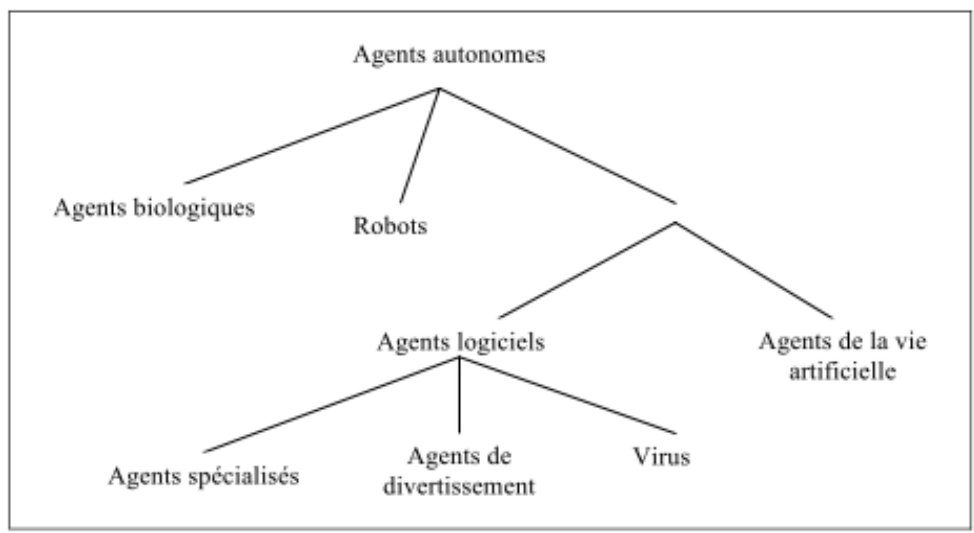

Figure 1 : La typologie des agents proposée par [**Franklin et Graesser , 96].**

- **[Rsseull et Norvig , 03]** [26] proposent une typologie des agents qui est présentée dans la figure ci-après :

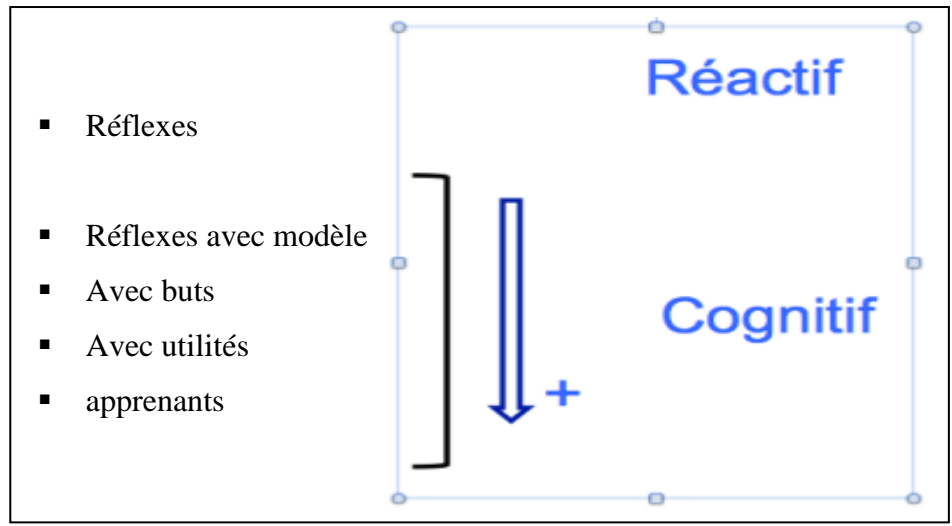

Figure 2 : La typologie des agents proposée par **[Rsseull et Norvig , 03].**

## 3. Systèmes mutlti_agents

### 3.1. Définition

La plupart des auteurs s'accordent généralement pour définir un système multi-agents (SMA) comme un système composé d'agents qui communiquent et collaborent pour achever des objectifs spécifiques personnels ou collectifs. La communication implique l'existence d'un espace partagé support de cette communication. Cet espace est généralement qualifié d'Environnement.

Pour [**Ferber, 95**] [10] : un Système Multi-Agents est un système composé des éléments,suivants :

- Un environnement E, c'est-à-dire un espace disposant généralement d'une métrique ;
- Un ensemble d'objets O. Ces objets sont situés, c'est-à-dire que pour tout objet, il est possible, à un moment donné, d'associer une position dans E. Ces objets sont passifs, c'est-à-dire qu'ils peuvent être perçus, créés, détruits et modifiés par les agents ;
- Un ensemble A d'agents qui sont des objets particuliers (A ⊆ O), lesquels représentent les entités actives du système ;
- Un ensemble de relations R qui unissent des objets (et donc des agents) entre eux ;
- Un ensemble d'opérations Op permettant aux agents de A de percevoir, produire, consommer, transformer, et manipuler des objets de O ;

- Des opérateurs chargés de représenter l'application de ces opérations et la réaction du monde à cette tentative de modification, que l'on appellera les lois de l'univers.

**3.2. Caractéristiques des systèmes multi-agents**

Selon [27] les SMA se caractérisent par :

3.2.1. **L'autonomie** : L'autonomie désigne communément l'indépendance et la capacité d'action et de prise de décision. Dans les SMA, l'autonomie peut se définir en trois points essentiels : l'existence propre et indépendante de l'agent, le maintien de sa viabilité en dehors de contrôle extérieur et la prise de décision en tenant compte uniquement de ses perceptions et de ses connaissances. En effet, les agents sont capables de planifier leurs actions, de raisonner et de résoudre des problèmes sans contrôle extérieur.

3.2.2. **La distribution** : La distribution dans les SMA est une caractéristique qui répond au besoin de distribution physique des connaissances et des traitements. Cette distribution impose un découpage modulaire dont la nécessité est parfois accentuée par l'absence de modèle global du problème à résoudre. Dans un environnement multi-agents, la distribution signifie donc que plusieurs agents participent à la réalisation d'un objectif global en se partageant les connaissances, les traitements, les tâches et les ressources.

3.2.3. **La décentralisation** : La décentralisation signifie la répartition du contrôle. L'éventuelle complexité du problème à résoudre rend difficile à l'utilisateur de gérer le contrôle total du système. Ainsi, une approche décentralisée de résolution consiste à léguer à chaque agent une partie de ce dernier. La décentralisation peut être dictée, entre autres, par des contraintes liées à la distribution physique du système ou par des limitations des capacités de décision globale.

3.2.4. **La communication** : La communication permet aux agents d'échanger des informations et assure ainsi la cohérence du comportement global du système malgré la décentralisation. Il existe plusieurs moyens de communication entre les agents :
  - ✓ **Communication indirecte via l'environnement :** Dans ce mode de communication les agents se communiquent par signaux via l'environnement. Ces signaux, une fois interprétée, vont produire des effets sur les agents. Ce type de communication est spécifique aux agents réactifs.

- ✓ **Communication par partage d'informations :** Les agents ne sont pas en liaison directe mais communiquent via une structure de données partagée, où on trouve les connaissances relatives à la résolution qui évolue durant le processus d'exécution. Elle suppose l'existence d'une base partagée sur laquelle les composants viennent lire et écrire. Cette manière de communiquer est l'une des plus utilisées dans la conception des systèmes multi-experts.
- ✓ **Communication par envoi de messages**: Les agents envoient leurs messages directement et explicitement au destinataire.
    - ➤ Mode point à point : l'agent émetteur du message connaît et précise l'adresse de ou des agents destinataires. Ce type de communication est généralement le plus employé par les agents cognitif.
    - ➤ Mode par diffusion : le message est envoyé à tous les agents du système. Ce type de transmission est très utilisé dans les systèmes dynamiques ainsi que les systèmes d'agent réactif.

Pour réaliser ce mode de communication, un langage de communication est nécessaire à travers lequel les agents arrivent à échanger des messages. Parmi ces langages en trouve : KQML et FIPA ACL.

- ➤ Le langage KQML : KQML est un langage permettant de structurer des messages afin de partager de l'information entre agents [28]. KQML est basé sur la théorie des actes de langage. Il propose une encapsulation des messages dans une performative qui définit l'acte illocutoire. Il est constitué d'un nombre important de performatifs qui sont en quelque sorte les opérations permises entre les agents. Les 36 performatifs de KQML peuvent être classifiés en trois grandes catégories :
    - o Les 18 performatives de discours : servent à échanger des connaissances et des informations présentes dans la base de connaissance de l'agent. Exemples: ask-if, ask-one, tell, describe, stream-all, etc.
    - o Les 11 performatives d'interconnexion : aide à la mise en relation des agents entre eux. Exemples: register, unregister, broadcast, etc.

- Les 7 performatives d'exception: servent à changer le déroulement normal des échanges. Exemples: error, sorry, standby, etc.
➢ Le langage FIPA ACL : FIPA-ACL [29] : Les performatifs de ce langage peuvent être regroupés en cinq groupes selon leurs fonctionnalités [30]:
- Actes pour le passage d'information : Inform, Inform-if, Inform-ref, Confirm, Disconfirm.
- Actes pour la réquisition d'information : Query-if, Query-ref, Subscribe.
- Actes pour la négociation : Accept-proposal, Cfp, Propose, Reject-proposal.
- Actes pour la distribution des tâches : Request, Request-when, Request-whenever, Agree, Cancel, Refuse.
- Actes pour le traitement des erreurs : Failure, Not-understood.

**3.2.5. L'interaction** : Les agents interagissent au sein de l'environnement dans lequel ils évoluent. Selon [10] « *Une interaction est une mise en relation dynamique de deux ou plusieurs agents par le biais d'un ensemble d'actions réciproques* ». Les interactions s'expriment ainsi à partir d'une série d'actions dont les conséquences exercent en retour une influence sur le comportement futur des agents. Les agents interagissent le long d'une suite d'événements pendant lesquels les agents sont d'une certaine manière en contact les uns avec les autres, que ce contact soit direct ou qu'il s'effectue par l'intermédiaire d'un autre agent ou de l'environnement. Sur la base des trois composantes principales de l'interaction, à savoir la nature des buts, l'accès aux ressources et les compétences des agents, plusieur situation d'interaction, entre les agents, sont possibles. Parmi ces situations :
- ✓ **L'antagonisme entre agents** : les agents ont des objectifs conflictuels (compétition) ou ont besoin de ressources communes (conflit sur les ressources).
- ✓ **L'indifférence entre agents** : les agents n'ont pas besoin les uns des autres pour atteindre leurs objectifs et ne sont pas gênés par ceux-ci.
- ✓ **La coopération entre agents** : les agents doivent s'entraider pour atteindre leurs objectifs qui peuvent être éventuellement communs.

- ✓ **La collaboration entre agents**: les agents interagissent entre eux afin d'accomplir un travail dont les tâches sont réparties entre l'ensemble de ces agents.
- ✓ **Négociation entre agents** : les agents interagissent sous un ensemble de règles de comportement pour limiter les conflits. La négociation consiste à trouver un compromis dans une situation conflictuelle ou changer les croyances de certains agents afin d'imposer l'avis d'un autre agent.

**3.2.6. L'organisation** : Il existe de multiples relations complexes qui unissent les agents et qui peuvent porter sur les buts, les plans, les actions ou les ressources. Ces relations induisent des schémas globaux d'interactions entre les agents. Les organisations permettent donc de formaliser ces schémas et offrent un moyen de spécifier et de concevoir une structure du SMA qui définit l'ensemble des rôles et des relations existant entre ces rôles. Les organisations constituent donc à la fois le support et la manière dont se passent ces interrelations, c'est-à-dire la façon dont sont réparties les tâches, les informations, les ressources, et la coordination d'actions. Selon [31], [32], [33] on distingue plusieurs types d'organisations :

- ✓ Groupe : plusieurs types de groupes existent :
  - ➢ Groupe simple: dès qu'un groupe existe, on peut avoir une coordination coopérative afin d'atteindre un but commun et partagé.
  - ➢ Équipe: une collection d'individus qui ont été rassemblés pour travailler ensemble. Dans une organisation plusieurs équipes sont formées pour des raisons de communication. Dans cette définition, des individus appartenant à une équipe doivent nécessairement communiquer entre eux, ce qui entraîne l'introduction de la notion d'environnement (cadre dans lequel les agents existent et évoluent).
  - ➢ Groupe d'intérêts: chaque membre a les mêmes intérêts, ils partagent les informations et coopèrent pour réaliser un but commun.
  - ➢ Communauté de pratique: se constitue lorsque des professionnels se regroupent et s'organisent pour partager des informations et des expériences relatives à leurs activités. Les membres de ces communautés peuvent ainsi échanger et coopérer afin de résoudre ensemble les problèmes auxquels ils peuvent être confrontés, apprendre ainsi les uns des autres et construire ensemble des connaissances et des pratiques communes.

- ✓ Hiérarchie où l'on distingue :
  - ➢ Hiérarchie simple: basée sur une relation maître/esclave, ce type d'organisation n'est plus utilisé.
  - ➢ Hiérarchie multi niveaux: les liens d'autorité forment un arbre. Le contrôle dans ce type d'organisation est très complexe, citons par exemple le problème d'allocation de ressources ou de planification qui doivent être pris en compte.
- ✓ Organisation décentralisée: c'est une hiérarchie multi divisions où chaque sommet d'une branche est une organisation à part entière. La difficulté principale dans ce type d'organisations est l'intégration des différents résultats provenant des différentes divisions.
- ✓ Marché: ce type d'organisation se base sur la relation consommateur/ producteur. Une instance spéciale du marché est le Contract Net Protocol qui est un protocole permettant l'élaboration et l'exécution d'un contrat entre un agent manager et un agent contractant. Il fait intervenir des agents interagissant entre eux pendant l'élaboration et l'exécution du contrat au moyen de performatifs.
- ✓ Coalitions : Une coalition est une organisation à court terme basée sur des engagements spécifiques et contextuels permettant aux agents de bénéficier de leurs compétences respectives de façon opportuniste.

**3.2.7. L'adaptation** : Un SMA est soumis à divers types de contraintes qui peuvent affecter ses performances. Il est donc essentiel que le système change son comportement lorsqu'il estime qu'il est entrain de dévier de son objectif global ou lorsqu'il s'avère qu'il peut réaliser une meilleure performance. L'adaptation est donc la capacité du système à modifier son comportement en cours de fonctionnement pour l'ajuster dans un milieu dynamique.

**3.2.8. L'ouverture** : Les SMA sont constitués de plusieurs entités autonomes, hétérogènes, en interaction entre elles au sein d'environnements dynamiques. Au-delà de cette hétérogénéité, les SMA sont caractérisés par l'ouverture qui se manifeste par l'évolution fonctionnelle du système. Cette évolution correspond à l'ajout, la modification ou la suppression dynamique d'entités du système.

**3.2.9. L'émergence** : L'émergence est l'apparition progressive de comportements non spécifiés a priori au sein du système. En effet, la fonction globale du système est attendue à partir des spécifications locales de chacun des agents, elle n'est pas programmée à l'avance et elle apparaît comme résultat des interactions des agents entre eux. Cette fonction survient sans organisateur extérieur du système.

**3.2.10. La situation dans un environnement** : L'environnement d'un SMA est vu comme étant un espace partagé par l'ensemble des agents. C'est le lieu commun au sein duquel les agents agissent et s'influencent les uns les autres. Les agents interagissent avec leur environnement par le biais des perceptions et des actions qu'ils peuvent effectuer sur lui. [Russell et Norvig, 03] [34] proposent plusieurs propriétés pour classifier les environnements des SMA:

- ✓ **L'accessibilité** : dans un environnement accessible les agents peuvent accéder à son état intégral. Par contre, la portée des actions de l'agent et de sa perception est locale dans un environnement inaccessible.
- ✓ **Le déterminisme** : l'état d'un environnement déterministe est lié seulement à son état précédent et à l'action. Dans un environnement indéterministe, les résultats de la même action et dans le même contexte peuvent être différents.
- ✓ **Le dynamisme** : un changement de l'état dans un environnement statique se produit exclusivement par les actions des agents. Cependant, l'état d'un environnement dynamique peut être modifié sans l'intervention des agents.
- ✓ **La continuité** : si le nombre de perceptions et d'actions possible de l'agent dans un environnement est illimité, on dit que l'environnement est continu. Sinon, on le considère discret.

**3.2.11. La délégation** : Dans une application multi-agents, l'utilisateur délègue son contrôle, ou du moins une partie de ce dernier, au système car il ne maîtrise pas le comportement de l'application globale. Cette délégation est due à la complexité de l'application et à l'incapacité de l'utilisateur à gérer toutes les décisions. L'utilisateur délègue son contrôle plus précisément aux agents, dont le caractère autonome et proactif leur permet de prendre les décisions à sa place.

**3.2.12. La personnalisation** : La personnalisation consiste à ce qu'un agent maintienne les préférences de l'utilisateur en observant ses comportements afin de construire un profil adapté à ce dernier. Ce profil est ensuite utilisé, par exemple, pour aider l'utilisateur à accéder à des informations pertinentes qui le concernent. Cette aide logicielle s'appelle "Profil Utilisateur".

**3.2.13. L'intelligibilité** : Un système intelligible est un système compréhensible et facilement abordable par les utilisateurs. Dans les SMA, l'intelligibilité découle du caractère anthropomorphique des agents. En effet, on retrouve souvent dans ces systèmes, une abstraction réaliste d'entités du monde réel à travers les agents.

## 3.3. Modélisation des systèmes multi-agents

Cette section présente trois tentatives d'extensions du modèle UML pour intégrer les concepts nécessaires à la modélisation orientée-agent. Elle détaille notamment les langages AUML, AML et AORML.

### 3.3.1. AUML (Agent Unified Modeling Language) :

AUML est une extension de UML [35], [36], [37], [38] dans laquelle divers diagrammes UML ont été étendus. En effet, [37] et [35] redéfinissent notamment les diagrammes de séquence pour permettre la spécification des protocoles d'interaction entre agents. Ils définissent la notion de rôle d'agent, de ligne de vie, ainsi que les différents types de protocoles. La sémantique des messages UML est également étendue pour intégrer notamment les actes de langage. [39] étend les diagrammes de classes et plus spécifiquement le concept de classe (représentant un agent) pour y intégrer les notions d'actions, de capacité et d'état de l'agent notamment utilisé pour représenter les croyances, les désirs et les intentions des agents (architecture BDI). Une partie est également consacrée à la représentation du comportement de l'agent sous forme d'automate à états. AUML est désormais en partie intégré dans certains outils tels que INGENIAS Development Kit [40] ou Opentool [41] associé à la méthodologie ADELFE.

### 3.3.2. AML (Agent Modeling Language)

[42] fournit tout un ensemble d'extensions à UML. AML définit trois types d'éléments semi-abstraits nommés semi-entités :

- Les semi-entités comportementales représentent des éléments possédant des capacités propres et capables d'agir, d'observer ou de percevoir leur environnement.
- Les semi-entités sociales qui peuvent former des sociétés, avoir des relations sociales et posséder des propriétés sociales propres.
- Enfin, les semi-entités mentales qui représentent des éléments pouvant être caractérisés en termes d'états mentaux : croyances, objectifs, besoins, désirs, buts, etc.

Ces semi-entités peuvent ensuite être surchargées par héritage pour intégrer les besoins d'approches plus spécifiques. Les entités fondamentales composantes d'un SMA telles les agents, les ressources et l'environnement disposent également de leurs définitions dans ce langage. AML peut lui-même être aisément étendu puisqu'il définit un métamodèle complet héritant de la super-structure UML.

### 3.3.3. AORML (Agent-Object-Relationship Modeling Language)

[43] Dans cette approche, deux points de vue différents sont adoptés sur la notion d'agent.

- Le point de vue externe décrit les agents, leurs interactions dans le domaine de l'application, leurs croyances sur les objets qu'ils manipulent ainsi que les relations qui relient ces différents éléments. Ces différents points sont représentés par les diagrammes d'Agent (Agent Diagram), de fenêtre d'interaction (Interaction Frame Diagram), de séquence d'interaction (Interaction Sequence Diagram) et de schéma d'interaction (Interaction Pattern Diagram). Le diagramme d'Agent décrit les différents types d'agents et les objets du domaine. Le diagramme de fenêtre d'interaction décrit les interactions possibles entre deux types d'agents, les types d'événements possibles ainsi que les types d'engagements. Le diagramme de séquence d'interaction décrit une instance d'un processus d'interaction. Enfin, le diagramme de schéma d'interaction décrit les schémas généraux d'interaction par un ensemble de règles de réactions définies dans le type de processus d'interaction.

- Le point de vue interne décrit quant à lui les différentes composantes internes d'un agent. Ce point de vue est associé aux diagrammes de fenêtre de réaction (Reaction Frame Diagram), de séquence de réaction (Reaction Sequence Diagram) et de schéma de réaction (Reaction Pattern Diagram). Le diagramme de fenêtre de réaction décrit les autres agents ou types d'agents, les types d'actions et d'événements ainsi que les engagements qui déterminent les diverses interactions possibles que l'agent en cours d'étude peut avoir avec eux. Le diagramme de séquence de réaction décrit les instances du processus d'interaction donné du point de vue interne de l'agent étudié. Enfin, le diagramme de schéma de réaction se concentre sur les modèles de réaction de l'agent en cours d'étude, ces réactions sont également exprimées sous la forme de règles de réaction.

## 3.4. Méthodologies orientés agent

Dans les dix dernières années, de nombreux efforts de l'ingénierie logicielle orientée agent se sont portés sur la définition de méthodologies pour guider le processus de développement des systèmes multi-agents. Les méthodologies orientés agent existantes constituent soit une extension des méthodologies orientées-objet, soit une extension des méthodologies à base de connaissances soit des méthodologies orientées organisation.

### 3.4.1. Les méthodologies constituant une extension des méthodologies orientées objet :

Les méthodologies orientées-objet sont populaires, en ce sens que plusieurs d'entre elles ont été utilisées avec succès dans l'industrie. L'expérience et le succès liés à cette utilisation peuvent faciliter l'intégration de la technologie agent. En effet la plupart des méthodologies existantes s'inspirent des résultats et des contributions issus de l'ingénierie logicielle orientée objet en y intégrant les spécificités liées à l'approche agent telles que l'autonomie ou la nature sociale des agents (PASSI [44], [45], MESSAGE [46], AAII [47], MaSE [14], ADELFE [49]). L'utilisation des méthodologies orientées objet a facilité le développement des méthodologies orientées agent, cependant une grande différence existe entre les deux méthodologies. En effet les méthodologies orientées objet modélisent le système en composants qui s'appellent objets. Elles se concentrent seulement sur la modélisation des composants dans le système en trouvant les attributs et les méthodes de chaque objet. Cependant les méthodologies orientées agent, au même titre que les méthodologies précédentes, se concentrent sur la modélisation des composants dans le système en utilisant des modèles : le modèle des agents, le modèle des rôles et le modèle des buts, Mais elle se concentrent aussi sur la modélisation des *interactions* entre les composants dans le système en utilisant des modèles des protocoles.

Parmi les méthodologies constituant une extension des méthodologies orientées-objet on trouve :

- **MaSE (Multiagent System Engineering) :** Cette méthodologie transforme les exigences entrées par l'utilisateur aux modèles qui décrivent le type d'agent, ses interactions avec les autres et l'architecture interne de chaque agent. Elle contient les étapes suivantes:

- ✓ **Captation des buts** : identifie et partitionne les buts du système. Puis, elle les organise dans un diagramme sous forme hiérarchique selon les relations entre les buts.
- ✓ **Description des cas d'utilisation** : extraire les scénarios qui représentent le comportement du système dans les cas spécifiques, ceci est à la base des exigences entrer par l'utilisateur et le modèle des buts. Chaque scénario décrit en détail les partenaires, l'ordre des activités et les informations échangées entre eux.
- ✓ **Construire l'ontologie** : construit la connaissance dans le domaine d'application du système en modélisant les informations échangées dans les scénarios.
- ✓ **Construire les rôles** : identifie les rôles du système ceci est à la base des diagrammes séquences et les buts. Chaque but est capté par au moins un rôle, et vis et versa, un rôle s'occupe au moins d'un but. Selon les diagrammes des scénarios, on détermine les relations nécessaires entre les rôles.
- ✓ **Représentation des tâches** : décrit en détail les tâches à réaliser pour obtenir le but. Chaque tâche est décrite par un diagramme des états dans lequel on indique l'état initial, l'état final et les états intermédiaires. Chaque état a des entrées, des sorties, des fonctions et des conditions provoquées et terminées.
- ✓ **Création des classes d'agent** : identifie les agents en se basant sur les rôles. Chaque rôle est joué par au moins un agent et chaque agent doit jouer au moins un rôle. Les relations entre les agents correspondent aux relations entre les rôles qu'ils jouent.
- ✓ **Construire des conversations** : les conversations sont des protocoles de coordination entre deux agents, cette étapes décrit en détail chaque conversation entre les agents. Pour chaque conversation, on détermine les participants, l'agent initial, l'agent final, les états. Chaque état a des fonctions, des conditions d'entrées, des conditions de sorties, des activités.

## 3.4.2. Les méthodologies constituant une extension des méthodologies à base de connaissance :

Les méthodologies à base de connaissance fournissent des techniques pouvant prendre en compte l'état mental des agents. De plus, ces méthodologies possèdent une librairie d'outils pouvant être utilisés. Cependant, ces méthodologies ne peuvent pas modéliser le comportement social des agents dans un SMA. Parmi ces méthodologies on trouve :

- **MASCommonKADS (Multiagent System–Knowledge Analysis and Development System):** Etendu de la méthodologie de génie connaissance. Cette méthodologie modélise un système par des étapes suivantes :
    - ✓ **Modélisation des agents** : détermine les caractéristiques de l'agent : Les capacités de raisonnement, les comportements (percevoir/s'agir), les services, les groupes et hiérarchie des agents.
    - ✓ **Modélisation des tâches** : détermine les tâches dont l'agent peut s'occuper : les buts, les partitions des composants et les méthodes de résoudre le problème qui lie à la tâche.
    - ✓ **Modélisation des coordinations** : déterminent les conversations entre les agents : les interactions, les protocoles et les capacités nécessaires.
    - ✓ **Modélisation de connaissance** : modélise la connaissance du domaine en utilisant le modèle d'expert. Modèle d'expert détermine les connaissances dont les agents ont besoin pour obtenir leurs objectifs.
    - ✓ **Modélisation d'organisation** : développe le modèle d'organisation en se basant sur le modèle agent. Ce modèle présente les relations statiques ou structurées entre les agents.

## 3.4.3. Les méthodologies orientées organisation

Les méthodologies ont évolué, depuis une vision initiale où le système était essentiellement centré sur l'agent et sur ses aspects individuels, vers une vision où il est désormais considéré comme une organisation dans laquelle les agents forment des groupes et des hiérarchies, et suivent des règles et des comportements spécifiques [50]. L'évolution des méthodologies GAIA [51], [52] et TROPOS [53], [54] en sont d'ailleurs les exemples les plus frappants. Les approches centrées-agent qui s'intéressent essentiellement aux actions individuelles des agents et où le SMA est conçu sur la base des états mentaux des agents (but, désir, croyance, intention, engagement, etc), seront distinguées des approches centrées-organisation ou organisationnelles. Les premières sont généralement dépendantes d'architectures spécifiques d'agent et de ce fait s'avèrent peu compatibles avec la nature ouverte et hétérogène des systèmes complexes. En revanche comme le décrivent [55], [56], les approches organisationnelles offrent de nombreux avantages par rapport aux approches centrées sur l'agent :

- Hétérogénéité des langages : Si chaque groupe (instance concrète d'organisation) est considéré comme un espace d'interaction à part entière, des moyens de communication spécifiques peuvent être mis en place dans chacun d'entre eux tels que KQML ou ACL, et ce, sans avoir à modifier l'architecture du système.
- Modularité : Chaque organisation peut être considérée comme un module qui décrit un comportement particulier pour ses membres. Chacune peut donc être utilisée pour définir des règles de visibilité claire au sein du processus de conception de SMA.
- De multiples architectures et applications : L'approche organisationnelle ne faisant aucune présupposition sur l'architecture interne des agents, elle laisse ouverte la spécification à un grand nombre de modèles et de techniques d'implantation.
- Sécurité des applications : Si tous les agents communiquent sans aucun contrôle extérieur, cela peut aisément entraîner des problèmes de sécurité. En revanche, si au sein de chaque groupe l'accès aux rôles peut être réglementé lorsque cela est nécessaire, on augmente alors le niveau de sécurité sans avoir à faire appel à un système de contrôle centralisé ou « global ».

Parmi ces méthodologies on trouve :

- **GAIA** : C'est d'aller de l'abstrait au concret; cette méthodologie exploite l'abstraction organisationnelle pour fournir une méthodologie d'analyse et de conception de systèmes de logiciel ouverts et complexes. Elle contient les modèles suivants:
    - ✓ **Modèle d'organisation** : divise le système en sous systèmes. Soit selon l'identification des sous systèmes qui existent déjà dans le système, soit selon la structure des composants. Ces derniers appartiennent au même sous système quand ils ont, soit, les objectifs communs, soit qu'ils interagissent avec une haute fréquence, soit que leurs capacités sont proches.

- ✓ **Modèle d'environnement**: l'environnement est considéré comme une liste de ressources. Chaque ressource étant associée avec un nom caractérisé par l'action où l'agent peut agir.
- ✓ **Modèle préliminaire de rôle** : ce n'est pas encore l'organisation actuelle. Il est une définition préliminaire des rôles et des protocoles de l'organisation. Gaia propose deux termes pour représenter de façon semiformelle le rôle, à savoir :
  - *Permission* : définit la relation de l'agent avec son environnement ; l'agent a t-il la permission d'accéder ou non à la ressource, de la changer ou de la consommer.
  - *Responsabilité* : détermine les comportements d'un agent selon deux types : 1/propriété vivant qui décrit les états auxquels un agent doit arriver et dans quelles conditions il doit arriver. 2/propriété sécurité qui assure qu'un agent arrive aux états acceptables. Le modèle préliminaire permet de sortir avec un ensemble des schémas de rôle.
- ✓ **Modèle préliminaire d'interaction** : définit les indépendances et les relations entre les rôles dans le système.
- ✓ **Modèle de règle organisationnelle** : la règle organisationnelle est considérée comme la responsabilité de l'organisation. Elle contient deux types : la règle vivante et la règle sécurité. La règle vivante assure l'ordre de réalisation des rôles ou des protocoles. La règle sécurité assure qu'un rôle est joué par au moins un agent et qu'un agent peut jouer au plus un rôle à la fois.

### 3.5. Implémentation des SMA

En conséquence de la multiplicité des méthodologies de conception, ces dernières années, plusieurs plates-formes orientées-agent ont été créées dont le but de permettre la mise en œuvre de modèles à base d'agents. En effet, les outils et logiciels issus de la technologie orientée-objet ne sont pas suffisamment adaptés au concept d'agent dans la mesure où ils ne permettent pas d'exprimer les caractéristiques d'autonomie, de proactivité, de réactivité des agents ainsi que les caractéristiques de socialité et de dynamique du SMA. Des plates-formes et outils de développement orientés-agents ont donc vu le jour pour offrir un support de mise en œuvre adéquat aux développeurs de SMA. Certains se basent sur des formalismes agent spécifiques et d'autres sont issus de la continuité de l'approche objet.

**[Arlabosse et al, 04]** [57] classent les plates-formes orientées-agent en différentes catégories dont:

### 3.5.1. Les approches orientées-langage
proposent de nouveaux langages orientés-agent ou plus généralement étendent des langages existants pour y intégrer le concept d'agent. Parmi ces langages :

- **Concurrent METATEM** [58] : est un langage de programmation multi-agents basée sur une sémantique très proche de la logique temporelle. Un système basé METATEM contient un certain nombre d'agents exécutés de manière concurrente, et communiquant par l'intermédiaire de message asynchrone (en "broadcast"). Le comportement de chaque agent est programmé à l'aide d'une spécification en logique temporelle. METATEM permet également de prouver que l'exécution de la spécification d'un agent est correcte et conforme à la spécification initiale.

- **Jack** [59], [60] : le langage agent Jack est un langage de programmation qui étend Java avec les concepts agents tel que agent, capacité, événements, plans et introduit les mécanismes de gestion des ressources et de l'exécution concurrente. L'architecture des agents JACK est basée sur l'approche BDI. Jack intègre un ensemble d'outils graphiques pour l'analyse et la programmation.

- **Dima** (Développement et Implantation de Systèmes Multi-Agents) [61], [62]: l'architecture d'agents DIMA propose de décomposer chaque agent en différents modules, chacun représentant les différents comportements d'un agent tels que la perception, la communication et la délibération. La représentation des agents est fondée sur des mécanismes déclaratifs ainsi le mécanisme de contrôle d'un agent peut être décrit par un automate et son interpréteur. La première version de DIMA était en Smalltalk-80, elle fut portée ensuite en JAVA. Une version répartie appelée DARX (Dima Agent Replication Extension) existe également.

- **Jason** [63] : est un interprèteur et un langage orienté-agent. C'est une extension du langage AgentSpeak [64]. Comme son prédécesseur, il est principalement dédié à la conception d'agents BDI. Il gère les communications à base d'actes de langage et peut s'interfacer avec d'autres plates-formes telles que Jade par exemple. Il fournit également une implantation du modèle organisationnel MOISE [65] (Rôles, Groupe, et Mission).

### 3.5.2. Plates-formes pour agents cognitifs, communiquants, web ou mobiles

- ✓ **Synergic** [66] : l'architecture des agents y est plutôt cognitive. Chaque agent dispose de croyances sur son environnement représentées par des accointances. Les agents ne raisonnent pas sur l'organisation du SMA et n'en ont aucune représentation. Synergic était développée en C et n'est plus maintenue depuis 1993.

- ✓ **OSACA** (Open System of Asynchronous Cognitive Agents) [67], [68] : résulte de la combinaison des agents avec les architectures orienté-objet de type CORBA. OSACA est un environnement qui gère les systèmes ouverts et fournit un langage pour créer aisément des agents (LAG). Les agents sont cognitifs et développés par clonage de l'agent générique puis dotés de compétences. Le langage de programmation était initialement Lisp puis C ou C++. Des extensions ont ensuite été proposées tel que SMAS pour la simulation de SMA complexes et OMAS pour l'exécution temps réel.

- ✓ **Mocah** (MOdélisation de la Coopération entre Agents Hétérogènes) [69], [70] : est un modèle de coopération permettant à plusieurs modèles de raisonnement hétérogènes de coopérer pour résoudre un problème commun. Chaque modèle de raisonnement est représenté par un agent particulier et manipule un type de connaissances particulier (le domaine) et un mode de raisonnement (les méthodes) approprié à ces connaissances. Chaque agent est caractérisé par un comportement coopératif. Cette plate-forme a été conçue pour traiter les problèmes de diagnostic de pannes électriques.

- ✓ **AgentBuilder** [71] : est une plate-forme commerciale développée en Java permettant de construire des agents intelligents. Le comportement d'un agent est décrit grâce au langage AGENT-0. L'architecture Agent est basée sur l'approche BDI. Les agents communiquent par messages KQML. AgentBuilder est composé d'un toolkit qui fournit un environnement intégré pour assister le processus de développement des agents et d'un système d'exécution qui fournit l'environnement d'exécution des agents.

- ✓ **Jade** (Java Agent DEvelopment framework) [72] : est un middleware développé en java. Il est conforme aux spécifications FIPA et gère donc le cycle de vie des agents, les services d'annuaire (pages blanches et jaunes), mais également la migration des agents. Il est basé sur une architecture peer-to-peer et permet la distribution des applications et la configuration dynamique et à distance des différents nœuds de déploiement.

### 3.5.3. Plates-formes de simulation

- ✓ **Swarm** [73] : est l'outil privilégié de la communauté américaine et des chercheurs en Vie Artificielle. Swarm est disponible en Java ou en Objective C. Swarm est une plate-forme destinée à la simulation des systèmes complexes adaptatifs. Un système Swarm est composé d'unités basiques appelées Swarm qui représentent une collection d'agents. Swarm gère la modélisation hiérarchique et holoniques des systèmes. Swarm fournit un ensemble de bibliothèques orientées-objet pour la construction, l'affichage, l'observation et le contrôle des simulations.

- ✓ **Cormas** [74] : est un environnement de programmation permettant la construction de modèles de simulation multi-agents. Il fut conçu en particulier pour simuler des écosystèmes ou des problèmes de gestion des ressources communes. Il est clairement inspiré de Swarm, et développé en SmallTalk. L'espace est géré par un automate cellulaire où chaque cellule peut contenir un autre automate. Cette plate-forme offre des outils pour définir les agents et leurs interactions, contrôler la dynamique globale de l'environnement et observer la simulation (liens d'accointance, proximité interactionnelle, etc).

- ✓ **Mobidyc** [75], [76] : est une plate-forme de simulation spécialisée pour les domaines de l'écologie, de la biologie et de l'environnement. Elle dispose d'outils pour la création et l'utilisation de modèles individus-centrés ainsi qu'un module dédié à l'étude statistique de plans d'expériences simulatoires raisonnés. Comme dans Cormas, l'environnement est considéré comme discret et il est géré par un automate cellulaire.

### 3.5.4. Plates-formes pour agent à base de composants

- ✓ **Maleva** (Modular Architecture for Living and EVolving Agents) [77] : est une plate-forme développée à base de composants en Borland Delphi. Dans cette plate-forme, un agent est considéré comme un composant, lequel est lui-même construit par composition de différents composants. Un mécanisme de communication synchrone/asynchrone entre ces composants est assuré par un gestionnaire de message dédié. Les communications inter-agents fonctionnent sur le même principe que celles inter-composants. Maleva comprend des outils pour la conception des composants de base et des agents par composition fonctionnelle et structurelle. Un ensemble d'outils pour simuler les agents dans des environnements topologiques est également proposé.

- ✓ **Comet** [78], [79] : tout comme Maleva, Comet est une plate-forme basée sur la notion de composant, mais le principe de communication est différent. Les composants communiquent ici par événements asynchrones. Une formalisation du modèle opérationnel de COMET a été réalisée à l'aide des réseaux de pétri et du langage Z. Une adaptation de la plate-forme DIMA intégrant les concepts de COMET a également été réalisée.

- ✓ **MASK** (Multi-Agent System Kernel) [80] : projet complet à part entière également, MASK est basée sur l'approche de Conception Voyelle (AEIO) [81] et vise à fournir un ensemble de bibliothèques d'Agents (Hybrides et Réactifs), de manipulation d'Environnement (E), d'Interaction (Langages à protocoles et Forces) (I) et d'Organisation(O) ainsi que les outils d'aide à la programmation. Pour chaque élément, MASK fournit des éditeurs qui permettent d'affiner le SMA de manière déclarative.

- ✓ **Volcano** [82], [83] : est une plate-forme basée sur la méthodologie de développement Voyelle (AEIO) [Demazeau, 1997]. Dans la continuité de MASK, Volcano couvre les différentes briques définies dans Voyelle. C'est une plate-forme orientée composant et elle utilise son propre langage de description d'architecture Madel (Multi-Agent DEscription Language) pour décrire les composants et leurs interactions.

### 3.5.5. Plates-formes complètes, génériques et généralistes

- ✓ **MACE** [84], [85] : fut l'un des premiers environnements généraux de modélisation pour SMA (indépendant du domaine d'application). Il introduit l'idée que les agents peuvent être utilisés dans tous les aspects de la construction de modèle (analyse et conception) et du développement. Dans MACE, un agent est un objet actif qui communique par envoi de messages. Un agent dispose de trois types d'actions : modifier son état interne, envoyer des messages aux autres agents ou envoyer des requêtes au noyau pour contrôler les événements internes. MACE utilise un système d'exécution concurrente et introduit déjà les concepts liés à la composition récursive d'agents : un groupe d'agents peut être considéré comme un agent. Les concepts liés à l'auto-organisation au sein d'un SMA sont également abordés.

- ✓ **Geamas** [86], [87], [88] : (Generic Architecture for MultiAgent Simulation) est une plate-forme logicielle générique pour la modélisation et la simulation multi-agents, implantée en Java. L'architecture logicielle de la plate-forme s'appuie sur un micro-noyau générique JAAFAAR offrant les structures et mécanismes minimaux nécessaires à l'implantation de SMA. A ce dernier, un certain nombre d'extensions logicielles spécialisées ont été adjointes tel que des modules d'apprentissage, d'auto-organisation ou de conception assistée, etc. Geamas gère trois niveaux d'abstraction : niveau micro (réactifs), médium (cognitifs, groupe) et macro (société). Gameas propose également un environnement intégré et des outils d'observation des applications.

- ✓ **Madkit** [89], [90], [91] : est une plate-forme basée sur le métamodèle organisationnel AALAADIN ou AGR. Elle est développée en Java et fournit un ensemble d'outils au développeur. Le cœur de Madkit est basé sur son micro-noyau qui fournit tous les services de base nécessaires aux agents : gestion des données organisationnelles (groupe, rôle), communication, ordonnancement (concurrent ou synchrone). Un ensemble de bibliothèques vient ensuite étendre les fonctionnalités de ce noyau pour notamment permettre la connexion de plusieurs noyaux et ainsi distribuer des applications, ou faciliter les simulations. L'une des originalités de Madkit est qu'il intègre les agents et ses principes de modélisation dans la conception même de la plate-forme. Madkit fournit également un environnement graphique qui permet de visualiser et de contrôler les agents. Le principal défaut de Madkit tient à son implantation du modèle organisationnel et notamment du concept de Rôle. Les rôles ne sont pas véritablement des comportements que les agents peuvent acquérir dynamiquement, mais sont réduits à de simples tags. Cette approche des rôles nuit gravement à la modularité et à la généricité des organisations.

- ✓ **MOCA** [92], [93] : est une plate-forme qui vient étendre Madkit et tenter de corriger ce défaut d'implantation des notions de rôle et d'organisation, mais vient également renforcer le contrôle d'accès à un rôle. Il intègre également en partie le formalisme OZS de description formelle des rôles proposé dans [94]. MOCA considère désormais l'organisation comme une véritable structure réutilisable. MOCA est basé sur un ensemble de composants Java. Un composant est une boîte noire qui fournit et requiert des compétences. Le rôle est désormais considéré comme un composant particulier dédié à l'interaction avec l'extérieur de l'agent. Quand un agent décide de jouer un nouveau rôle cela revient à lui ajouter un nouveau composant.

### 3.5.6. Plates-formes intégrant la vision holonique

- ✓ **Janus** [95] : est une plate-forme spécifiquement conçue pour implanter et déployer les systèmes multi-agents holoniques. Elle gère de manière native l'intégralité du modèle organisationnel et fournit l'ensemble des moyens nécessaires pour implanter les concepts du métamodèle CRIO (Capacité, Rôle, Interaction et Organisation).

Parmi toutes ces plates-formes, on peut tout de même constater l'émergence de la plate-forme JADE. En effet, JADE est aujourd'hui considérée comme la référence européenne en terme de satisfaction du standard FIPA et de nombreux projets ont été développés sur sa base ou ont contribué à ses extensions.

### 3.6. Domaines d'application des SMA

Les applications des systèmes multi-agents couvrent de plus en plus de domaines. Citons les systèmes d'information coopératifs, la simulation sociologique, les outils documentaires adaptés au Web, les robots autonomes coopératifs, jeux vidéo (multi-joueurs), résolution distribuée de problèmes, etc. Néanmoins les systèmes multi-agents développés actuellement peuvent être classés en trois catégories [96], [97], [98] :

- Les simulations dont l'objectif est la modélisation de phénomènes du monde réel, afin d'observer, de comprendre et d'expliquer leur comportement et leur évolution. Les systèmes multi-agents ont trouvé rapidement un champ extrêmement propice à leur développement dans le domaine de la modélisation de systèmes complexes ne trouvant pas de formalisation mathématique adaptée. Dans le domaine des sciences du vivant d'abord, ensuite dans celui des sciences humaines et sociales, les SMA ont montré qu'il était possible de modéliser au niveau micro les comportements d'entités élémentaires et d'étudier au niveau macro le résultat global de l'interaction de ces entités. En effet, la simulation multi-agents permet de tester rapidement le changement de certaines hypothèses ; elle permet aussi d'intégrer de nouveaux agents et d'éditer, sur un plan pratique, les résultats pour comparer les expérimentations les unes aux autres ; de plus elle permet de préserver l'hétérogénéité du système à simuler.

- Les applications dans lesquelles les agents jouent le rôle d'êtres humains. La notion d'agent simplifie la conception de ces systèmes et amène de nouvelles problématiques centrées utilisateur telles que la communication, la sécurité… Les systèmes de ventes aux enchères dans laquelle les agents jouent les rôles decommissaire priseur et d'acheteurs, représentent une classe d'applications de cette catégorie.

- La résolution de problèmes : telle qu'elle avait été définie en Intelligence Artificielle, étendue à un contexte distribué. Dans ce cadre, l'objectif est de mettre en œuvre un ensemble de techniques pour que des agents, pertinents pour la résolution d'une partie ou l'ensemble du problème, participent de manière efficace et cohérente à la résolution du problème global.

## 4. Conclusion

La synthèse de l'état de l'art sur les systèmes multi-agents présentée dans ce chapitre montre que la technologie orientée-agent est devenue un paradigme à part entière du génie logiciel disposant de ses propres éléments méthodologiques en termes de conception et de programmation. Le succès de ce paradigme, par rapport aux autres paradigmes, repose sur ses caractéristiques fonctionnelles et comportementales propres telles que l'autonomie, la proactivité, la sensibilité, la flexibilité, etc. Ces caractéristiques ont certes, largement contribué au grand succès de ce paradigme mais, elles ont introduit de nouveaux défis lors de l'application de ce dernier dans les différents domaines tel que la planification.

# Chapitre 2
# Planification multi-agents

1. **Introduction**

La planification [99], [100], [101], [102] en intelligence artificielle s'est intéressée depuis les années 1960 à la génération de plans (séquences d'actions) visant à atteindre un objectif fixé pour des problèmes exprimés dans un langage concis d'opérateurs de transformation d'état, souvent proche de la logique propositionnelle. La planification est donc le sous-domaine de l'intelligence artificielle qui cherche à répondre à la question: "Que doit-on faire?", c'est-à-dire quelle action poser et dans quel ordre. Ce chapitre présente une synthèse de l'état de l'art sur la planification. Il se compose de quatre parties. La première partie concerne la planification classique à savoir : sa définition, ses représentations, ses algorithmes. La deuxième partie concerne la planification hiérarchique. La troisième partie concerne la planification dans l'incertain à savoir : sa définition, ses représentations, ses algorithmes. La quatrième partie concerne la planification multi-agents à savoir : sa définition, les types de planification multi_agents, la coordination des plans.

2. **Planification classique**

Les recherches, qui ont menées à la planification actuelle, ont commencé dans les années 60, sous la forme de programmes dont le but était de simuler la capacité de raisonnement de l'être humain. A cette période on parlait de planification classique. Un des premiers programmes fut le General Program Solver (GPS) [103]. Le GPS fonctionne par recherche dans un espace d'états. Cette recherche est guidée par le calcul des différences entre l'état courant et l'état final du problème. A la fin des années 90 le domaine de la planification classique a connu un essor considérable; tant pour la richesse de modélisation des problèmes de planification que pour l'efficacité des systèmes de génération de plans.

### 2.1. Représentation classique de la planification

En pratique, il est impossible d'énumérer tous les états et les transitions entre états possibles. En effet, la description d'un problème peut être excessivement longue, et peut nécessiter plus de travail que la résolution du problème. Pour résoudre cette difficulté, il faut un langage de représentation du problème permettant de calculer ces états et ces transitions entre états à la volée. Dans la littérature, on recense trois façons de représenter les problèmes de planification classique [104] :

- la représentation dans la théorie des ensembles (*Set-theoretic representation*) : chaque état du monde est un ensemble de propositions et chaque action est une expression spécifiant les propositions qui doivent appartenir à l'état courant pour que l'action puisse être exécutée, ainsi que celles qui seront ajoutées et enlevées de l'état suite à l'exécution de l'action ;
- la représentation classique (*Classical representation*) : contrairement à la représentation précédente, des prédicats du premier ordre et des connecteurs logiques sont utilisés à la place des propositions ;
- la représentation par variables d'états (*State variable representation*) : chaque état est représenté par un tuple de n variables d'états valuées {x1, x2, ..., xn } et chaque action est représentée par une fonction partielle qui permet de passer d'un tuple à un autre tuple de variables d'états instanciées.

Les trois représentations ci-dessus décrites utilisent un formalisme étant très restreint, des extensions sont nécessaires afin de décrire des domaines plus complexes. Ces principales extensions sont le typage des variables, les opérateurs de planification conditionnelle, les expressions de quantification, les pré-conditions disjonctives ou encore les axiomes d'inférences. Le langage de planification PDDL (Planning Domain Description Language) [105] permet d'exprimer ces différentes extensions. Ce langage a permit de fédérer les recherches en planification classique, et grandement facilité la conception et le partage des meilleures techniques. Depuis sa création, de nombreuses évolutions ont été proposées :

- PDDL 1.2 [105] : cette version est inspirée du langage utilisé par le planificateur UCPOP. Elle définit la représentation de base des opérateurs (préconditions, effets) et des problèmes (état initial, but).
- PDDL 2.1 [107] : cette version introduit deux extensions majeures : la prise en compte du temps, et les variables numériques.
- PDDL 2.2 [108] : cette version introduit deux extensions mineures : les prédicats dérivés et les littéraux temporels initiaux.
- PDDL 3.0 [109] : cette version introduit deux extensions importantes : les contraintes sur les trajectoires et les préférences sur les buts.
- PDDL 3.1 [110] : cette version introduit les variables d'état binaires.

**2.2. Algorithmes de la planification classique**

De nombreux algorithmes ont été développés dans le cadre de la planification classique. Ils se rangent en deux catégories : (i) les algorithmes de génération des plans dont l'objectif est de trouver un plan optimal pour un critère donné, d'optimiser ce critère sans prouver l'optimalité, de trouver un plan rapidement sans optimisation….(ii) les algorithmes d'analyse structurelle de problème de planification qui consiste à analyser le problème de planification dont l'objectif est de guider ou contraindre la recherche d'une solution.

**2.2.1. Les algorithmes de génération des plans**

**a. Les algorithmes de recherche heuristique**

Les premiers types d'algorithmes parmi les plus employés sont les algorithmes de recherche heuristique [111], [112] [113] [114] . Ils sont utilisés pour effectuer une recherche dans les espaces d'états en chaînage avant ou arrière, ou bien dans les espaces de plans où un nœud représente un plan partiel et un arc représente une opération de modification d'un plan partiel (introduction d'une action, relation de précédence entre actions, protection d'un lien causal). Ces algorithmes sont utilisés soit pour la planification non optimale, soit pour la planification optimale en nombre d'actions ou en coût total. Parmi les planificateurs de recherche heuristique, on peut citer les planificateurs suivant :

- Le planificateur RePOP [115] et VHPOP [116] utilisent cette heuristique dans les espaces de plans partiels.
- le planificateur FF [114] calcule à chaque état la longueur d'un plan pour le problème relaxé .

- le planificateur YAHSP [117] ajoutant au précédent la construction gloutonne d'un plan anticipé dont l'état résultant est ajouté à la liste des nœuds à développer.
- le planificateur Metric-FF [118] et SAPA [119] pour la planification avec variables numériques.
- le planificateur SGPLAN [120] qui partitionne un problème en plusieurs sous problèmes plus simples.
- le planificateur CRICKEY [121] et COLIN [122] pour la planification temporellement expressive.
- le planificateur Macro-FF [123] et Marvin [124] qui calculent des macros-opérateurs ;
- le planificateur TFD [125] qui calcule une heuristique basée sur les graphes de transition de domaines.
- le planificateur LAMA [126] qui utilise une heuristique basée sur les landmarks.
- Le planificateur GAMER [127] pour la planification optimale en nombre d'actions ou en coût. Il utilise aussi des techniques de model-cheking comme les BDDs, ou des variations de TFD avec des heuristiques améliorées [128][129].
- le planificateur TP4 [130] pour la planification temporelle optimale.
- l'algorithme STRIPS [131] : l'algorithme strips fonctionne de la même façon que l'algorithme de recherche par recherche arrière, mais son avantage réside dans sa capacité à réduire l'espace de recherche. Cette optimisation est caractérisée par les deux points suivants :
  - à chaque appel récursif de l'algorithme, les sous-buts qui doivent être satisfaits sont les pré-conditions du dernier opérateur ajouté dans le plan, ce qui a comme conséquence de réduire substantiellement le facteur de branchement de l'algorithme ;
  - si l'état courant satisfait toutes les pré-conditions d'un opérateur, alors strips « marque » l'opérateur. Ainsi, en cas d'échec, le retour arrière se fera à partir de cet opérateur.

**b. Les algorithmes à base de contraintes**

Les approches à base de contraintes, comme la programmation par contraintes (PPC) ou la satisfaction de bases de clauses (SAT) sont très utilisées en planification. Parmi les planificateurs à base de contraintes, on peut citer les planificateurs suivant :

- Les planificateurs SAT-PLAN [132] et BLACKBOX [133] encodent un problème sous forme d'une base de clauses pour un horizon donné et utilisent ensuite un prouveur SAT pour trouver une solution [134]. Des améliorations des codages et planificateurs SAT ont récemment été proposées [135] , [136] Ces planificateurs calculent des plans parallèles optimaux, cas particulier de planification temporelle avec durées uniformes.

- Le planificateur CFDP [137] utilise la PPC sur une structure inspirée du graphe de planification avec des règles d'élagage adaptées à la planification séquentielle optimale.
- Le planificateur GP-CSP [138] encode un problème sous forme de CSP et calcule un plan parallèle optimal.
- Le planificateur CPT [139] est un planificateur temporel optimal qui encode un problème en CSP suivant un schéma basé sur la planification dans les espaces de plans et introduit de nombreuses règles d'élagage basées notamment sur les actions qui ne font pas encore partie d'un plan partiel.
- Le planificateur TLP-GP [134] qui effectue une recherche dans une structure inspirée du graphe de planification encodant des contraintes temporelles, pour la planification temporellement expressive.

**c. Les algorithmes utilisant d'autres techniques**

Dans ce type d'algorithme nous allons citer des algorithmes de planification basés sur d'autres techniques telles que les réseaux de Petri pour la planification temporelle optimale [140], les automates pondérés pour la planification factorisée optimale en coût [141], les algorithmes évolutionnaires [142] permettant d'optimiser la qualité des plans produits par un planificateur.

**2.2.2. Les algorithmes d'analyse structurelle de problèmes de planification**

L'analyse structurelle de problème de planification consiste à extraire automatiquement des informations à partir de la description d'un problème, permettant de guider ou contraindre la recherche d'une solution vers les portions de l'espace de recherche les plus prometteuses [143]. Parmi les techniques d'analyse de problème de planification on peut citer IPP [143] et STAN [144].

## 3. Planification hiérarchique

La planification hiérarchique (Hierarchical Task Network, htn ) [145] utilise un langage similaire au langage de la planification classique. En effet, chaque état du monde est représenté par un ensemble de prédicats instanciés et chaque action correspond à une transition déterministe entre deux états. Cependant la planification hiérarchique diffère de la planification classique sur la façon de planifier. Dans cette planification, l'objectif n'est pas d'atteindre un ensemble de buts mais de résoudre un ensemble de tâches. Les entrées du planificateur sont, comme en planification classique, un ensemble d'opérateurs, mais également, un ensemble de méthodes. Une méthode décrit la manière de décomposer une tâche en un ensemble de sous-tâches. Le principe de la planification hiérarchique est de décomposer les tâches non primitives en sous-tâches de plus en plus petites, jusqu'à obtenir des tâches primitives qui peuvent être résolues en utilisant les opérateurs de planification.

## 4. Planification dans l'incertain

La planification, dans sa version classique a connu un essor considérable à cause de la richesse des langages de modélisation et l'efficacité des systèmes de génération des plans. Néanmoins, la planification classique souffrait d'une faiblesse causée par le fait qu'elle reposait sur deux hypothèses simplificatrices fortes à savoir : la disposition d'une connaissance parfaite, à tout instant, de l'état du système et des effets des actions et la certitude que les modifications de l'état du système proviennent uniquement de l'exécution des actions du plan. Pour pallier à cette faiblesse, le domaine de la planification dans l'incertain s'est développé, proposant d'intégrer des actions à effet probabiliste puis des fonctions d'utilité additives sur les buts, conduisant à une famille d'approches pour la planification basées sur la théorie de la décision [146].

### 4.1. Représentation de la planification dans l'incertain

Au départ le processus décisionnels de Markov (PDM) a servi de base sémantique à certaines des approches proposées pour la planification dans l'incertain. Il a été l'approche privilégiée pour la représentation et la résolution de problèmes de planification dans l'incertain en intelligence artificielle. Cependant le cadre classique des PDM repose sur une représentation explicite des états, c'est-à-dire où chaque état est énuméré dans une table [147]. Cette modélisation présente deux inconvénients non négligeables pour des applications réalistes : les états sont souvent décrits par un ensemble de variables à domaines discrets ou continus, et, à cause de l'explosion combinatoire due au nombre de combinaisons possibles des valeurs des variables, les états ne peuvent pas être énumérés avant la résolution du problème. De nombreux travaux récents ont visé à améliorer ce cadre en le dotant du pouvoir expressif des langages de représentation traditionnellement utilisés en intelligence artificielle : logique, contraintes ou réseaux bayésiens. Cette amélioration a abouti aux modèles intensionnels de PDM. Ces modèles sont dits intensionnels, car ni les états ni les transitions ni les récompenses ne sont construits a priori, mais sont définis par des formules logiques portant sur les variables d'état du modèle. Chaque formule peut être instanciée en remplaçant les variables d'état par leurs valeurs à un instant donné, ce qui permet de construire au besoin l'état correspondant à ces variables instanciées, ainsi que les transitions ou les récompenses définies sur cet état. Plusieurs modèles intensionnels de PDM ont été proposé, les deux principaux modèles intensionnels de PDM sont: les réseaux bayésiens dynamiques [148], [149] et les modèles STRIPS probabilistes [150], [151].

### 4.2. Algorithmes de la planification dans l'incertain

De nombreux algorithmes ont été développés pour résoudre des PDM intensionnels. Ils se rangent en deux catégories fondamentalement différentes au niveau des méthodes utilisées : les approches probabilistes [152], [152], [153], [154], [149], [151], [155], [156], [157], qui étendent les méthodes de résolution des PDM classiques aux PDM intensionnels. les approches déterministes [158], [159], [160], [161], qui réduisent la résolution du PDM intensionnel à plusieurs résolutions de problèmes de planification classique déterministe.

### 5. Planification multi-agents

L'extension de la planification dans le cadre des systèmes multi-agent a abouti à la planification distribuée [4] [11] dans laquelle le domaine de planification est réparti sur un ensemble d'agents. Ces agents peuvent être coopératifs dans le sens où ils ont un objectif global commun et des capacités complémentaires pour le réaliser ou individualistes dans le sens où ils ont des objectifs individuels dont ils sont capables d'assurer la réalisation sans aide externe. Dans les deux cas les agents doivent être capables de générer des plans qui permettent la réalisation soit des sous-objectifs nécessaires pour un objectif global soit des objectifs individuels.

## 5.1. Types de la planification multi-agents

La planification multi-agents peut être divisée en trois types : **(i)** la distribution concernant uniquement la phase de l'exécution : selon ce type la planification est centralisée afin de produire un plan collectif cohérent mais l'exécution est distribuée c. a. d. différents agents exécutent chacun différentes actions du plan collectif, **(ii)** la distribution concernant uniquement la phase de planification : selon ce type différents agents génèrent et coordonnent des plans de manière distribuée afin de générer un plan collectif cohérent mais l'exécution est centralisée c.a.d. un agent peut exécuter le plan collectif, **(iii)** la distribution concernant les deux phases à la fois : selon ce type, la planification et l'exécution sont distribuées c.a.d. les différents agents génèrent et coordonnent des plans individuels de manière distribuée afin de générer un plan collectif cohérent et ensuite chaque agent exécute l'action du plan collectif qui correspond a ses capacités. Les trois types de planification multi-agents sont développés dans ce qui suit.

### 5.1.1. Planification centralisée pour des plans distribués (PCPD) [un planificateur, plusieurs exécutants]: se focalise sur le contrôle et la coordination d'actions réalisées par plusieurs agents dans des environnements partagés (figure3).

- **Principe**

  - Génération des plans à ordre partiel avec des contraintes d'ordonnancement minimales,
  - Décomposer le plan en sous-plans,
  - Insérer des actions de synchronisation dans les sous-plans pour assurer l'ordonnancement entre les agents,
  - Allouer les sous-plans aux agents,
  - Initier l'exécution du plan.

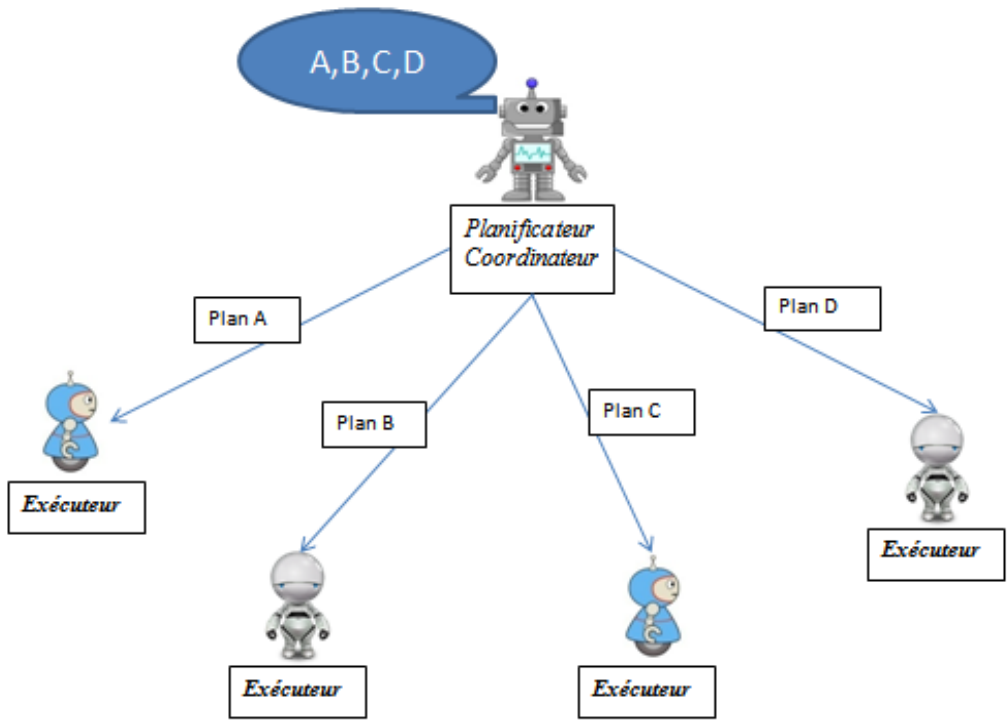

Figure 3 : Planification centralisée pour des plans distribués (PCPD).

- **Avantage**

    - Facilite la résolution de conflits et la convergence vers une solution globale.

- **Problème**

    - Nécessite la centralisation du contrôle au niveau d'un seul agent.

**5.1.2. Planification distribuée pour les plans centralisés (PDPC)** [plusieurs planificateurs, un seul exécutant]: se focalise sur le processus de planification et son extension à un environnement distribués (figure 4).

- **Principe**

    - Le processus de planification complexe est distribuée entre différents agents spécialistes qui coopèrent et communiquent en partageant des objectifs et des représentations,
    - L'objectif principal est de former un plan cohérent,

- Les spécialistes génèrent leurs portions de sous-plans,
- Si l'un des spécialistes ne satisfait pas, un retour arrière est réalisé en choisissant un nouveau partenaire,
- Les plans partiels des agents sont ensuite synchronisés en un seul plan (partage de résultats).

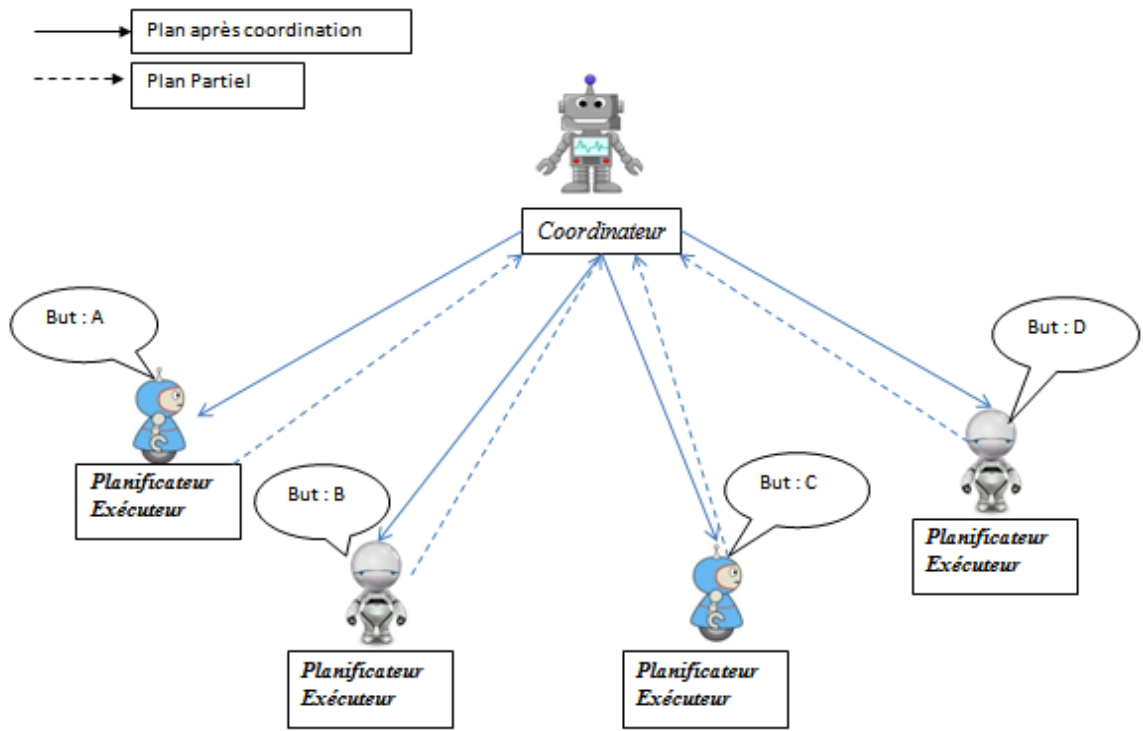

Figure 4 : Planification distribuée pour les plans centralisés (PDPC).

- **Avantage**

   - Pas de centralisation du contrôle.

- **Problème**

   - Cout de communication élevé,
   - Incompatibilité des buts et des intentions,
   - Connaissances incohérentes,
   - Représentation différente des plans partiels.

**5.1.3. Planification distribuée pour des plans distribués (PDPD)** [Plusieurs planificateurs, plusieurs exécutants]**:** se focalise sur le fais que chaque agent est capable de produire ses propre plans indépendamment (Agent-driven) ou en étant dirigé par un but commun avec les autre (goal-driven) (figure 5).

- **Principe**

    - le processus de synthèse et de d'exécution d'un plan multi-agent sont distribués. Ce type de planification peut être :
    - **Orientée tache :**
        - Existence d'un but globale
        - La synthèse des sous-plans porte sur les résultats et non les plans
    - **Orientée agent**
        - Pas de plan et de but global
        - Mécanisme permettant l'exécution de plans concurrents.

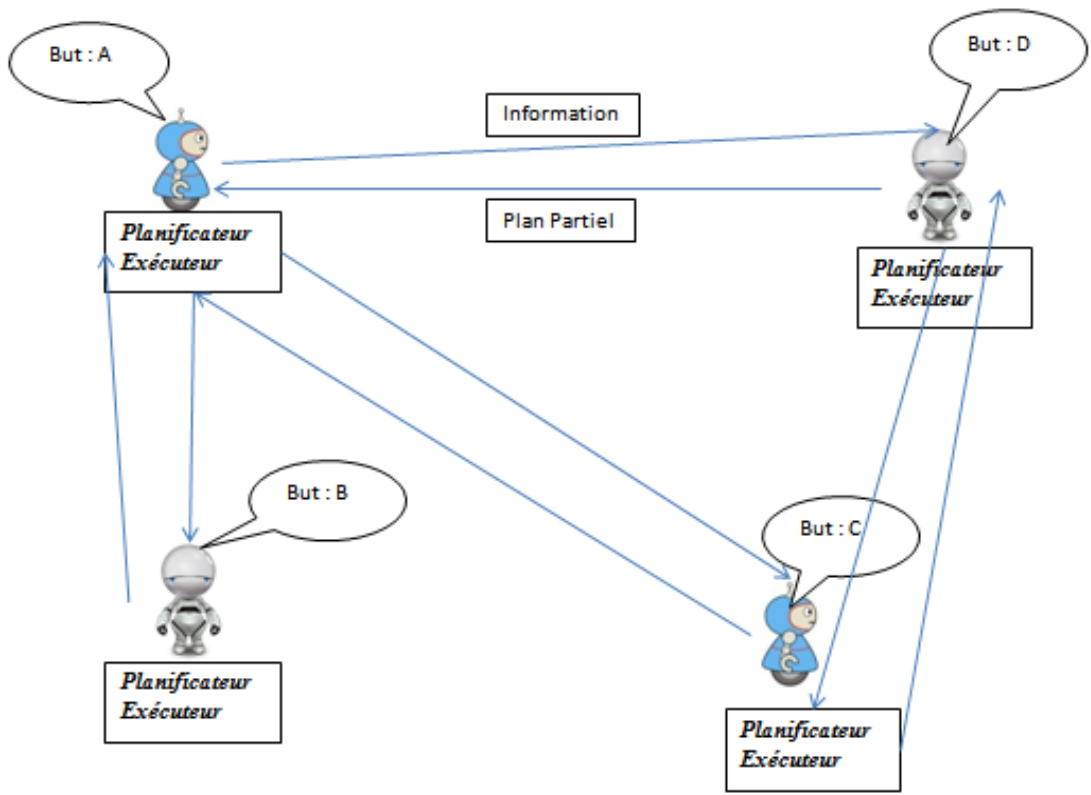

Figure 5 : Planification distribuée pour des plans distribués (PDPD).

- **Avantage**

    - Pas de centralisation du contrôle.
    - Le cout de la communication et minimal

- **Problème**

    - Absence de la cohérence entre les agents lors de l'exécution de leur plan
    - Un problème de coordination des plans causés par le partage des ressources

### 5.2. La coordination des plans

Le fait que la planification multi-agent soit distribuée sur un ensemble d'agents qui ont une forte dépendance entre leurs taches et qui partageant les même ressources, cette dernière doit s'intéresser, aussi, à la coordination entre ces agents lors de la génération des plans. Ceci à travers l'identification des relations entre les activités des agents dans le but d'éviter les conflits lors de l'exécution de ces plans. Le problème de la planification multi-agents consiste donc, à trouver un plan pour chaque agent qui réalise ses buts privés, de façon à ce que l'ensemble des plans soit coordonné et que les buts globaux soient atteints.

Comme évoqué précédemment, dans la planification multi-agents, les plans peuvent être construits de manière centralisée puis distribués aux agents, ou bien chaque agent peut construire localement son propre plan puis le coordonner de manière distribuée. Dans le premier cas, seule l'exécution du plan est distribuée. En revanche, dans le second, la génération des plans, le processus de coordination ainsi que l'exécution sont réalisés de manière distribuée. Nous présentons dans ce qui suit les mécanismes de coordination centralisée de plans, puis les quatre grands mécanismes de coordination décentralisée qui sont : la planification partiellement globale, la synchronisation distribuée de plans, la fusion incrémentale de plans et la coordination lors de l'exécution des plans.

- **Coordination centralisée des plans** : La coordination par planification centralisée repose toujours sur l'existence d'un agent coordinateur. Cet agent centralise l'ensemble des plans des agents du système et résout les conflits potentiels entre leurs activités en introduisant des actions de synchronisation. L'agent coordinateur peut : soit planifier pour l'ensemble des agents et, dans ce cas, il doit décomposer le plan global en sous-plans synchronisés pouvant être exécutés par les agents ;

soit chaque agent peut planifier localement et, dans ce cas, le rôle de l'agent coordinateur se limite à la synchronisation des plans reçus. Ces deux approches peuvent être résumées ainsi : **(i) approche par synchronisation** : la synchronisation de plans s'appuie sur la notion de plan non-linéaire. En effet, un plan non-linéaire [10] est un plan dont les actions ne sont pas strictement ordonnées. Par conséquent, il est possible de contraindre les relations d'ordre entre les actions en ajoutant des actions de synchronisation et garantir ainsi que le plan sera exempt de conflits. **(ii) approche hiérarchique** : Corkill [11] propose une implémentation distribuée de NOAH (Nets Of Action Hierarchies) [12]. Cette implémentation est fondée sur l'allocation de sous-buts d'un certain niveau à des agents. Les plans sont synchronisés niveau par niveau, c'est-à-dire que les agents négocient et résolvent les conflits de chaque niveau avant de raffiner le plan à un niveau inférieur.

- **Planification partiellement globale** : La planification partiellement globale, PGP (Partial Global Planning) [13] est un schéma générique de coordination distribuée pour la résolution de problèmes qui s'appuie sur une architecture de type tableau noir (zone de mémoire commune et partagée entre les agents). Ce schéma de coordination ne vise pas à résoudre des conflits entre les plans des agents, mais à permettre un gain de performance en terme de temps de calcul (en réduisant l'inactivité de certains agents et en minimisant les tâches redondantes). Chaque agent est capable de percevoir les modifications de la zone du tableau dont il est responsable, et d'interagir avec les autres pour confronter ces connaissances. Les interactions entre agents reposent sur la diffusion d'une structure de données, appelée plan partiel global.

- **Synchronisation distribuée de plans** : Les travaux basés sur la synchronisation voient les activités des agents comme des processus concurrents qui doivent être synchronisés. Dans ce cadre, [14] proposent un algorithme distribué qui permet de réaliser une telle coordination. Ces travaux s'appuient sur une représentation du monde par états de manière similaire au formalisme strips [15]. La production des plans est réalisée par les agents en fonction de leurs buts locaux. Chaque agent est alors responsable de la synchronisation de ses plans avec les autres agents. Pour qualifier un plan local validé dans un contexte multi-agent, i.e., exempt de conflits, [14] introduit la notion de « plan structuré ». La construction d'un plan structuré se décompose en deux phases. Tout d'abord, l'agent élabore un plan de manière indépendante sans tenir compte des plans produits par les autres agents. Puis, il essaie de synchroniser son plan en le diffusant aux autres agents.

- **Fusion incrémentale de plans**. L'insertion incrémentale de plans PMP (Plan-Merging Paradigm) [16] permet aux agents de planifier, de coordonner leurs plans et de les exécuter d'une manière distribuée et incrémentale sans être contraints de suspendre l'exécution des plans précédemment coordonnés. PMP permet à un grand nombre d'agents de concilier leurs plans individuels, afin d'éliminer tout risque de conflit durant la phase d'exécution. La validation d'un plan individuel est réalisée par insertion incrémentale dans un graphe de contraintes établi à partir des différents plans en cours d'exécution. Cette insertion transforme le plan individuel en un plan coordonné exécutable, tout en imposant des contraintes d'ordre entre les plans en cours d'exécution et le plan à valider. Les différents mécanismes présentés sont des mécanismes de coordination et de coopération dans le cadre des systèmes multi-agent. Ils permettent de garantir l'intégrité fonctionnelle du système, i.e., que les activités des agents soient exemptes de conflits, mais ils n'ont pas pour objectif de produire un plan global.

- **Coordination par planification itérative** : consiste à exécuter les plans partiels, et lors de l'exécution les points de conflits sont reconnus et gérés par les solutions possibles suivantes :

    - **Usage de la force d'une autorité** : d'un poids supérieur (ascendant d'un agent sur un autre).
    - **Arbitrage :** existence d'un agent arbitre ou médiateur qui dispose des différents points de vue et tente de résoudre le conflit.
    - **Négociation :** La négociation joue un rôle fondamental dans les activités de coopération en permettant aux agents en conflit d'entrent dans une série de tractations, d'échanges et de compromis de manière à parvenir à un accord, c'est-à-dire une solution qui satisfasse toutes les parties. Deux grandes catégories de négociation.

        ✓ Négociation par compromis

            o Chacune des parties relache les contraintes les moins importantes.

- Il y a accord lorsque toutes les contraintes sontsatisfaites.

✓ Négociation intégrante

- Cherche à identifier les buts profonds (changementde but).
- Trouver une solution qui satisfasse complètement ces buts et non les propositions de "surface".

Différentes protocole de négociation ont été développées en s'appuyant sur la riche diversité des négociations humaines dans divers contextes [168], [169], [170], [171] :

✓ Le protocole du réseau contractuel (Contract-Net) a été une des approches les plus utilisées pour les SMA [172]. Les agents coordonnent leurs activités grâce à l'établissement de contrats pour atteindre des buts spécifiques. Un agent, agissant comme un gestionnaire (manager) décompose son contrat (une tâche ou un problème) en sous-contrats qui pourront être traités par des agents contractants potentiels. Le gestionnaire annonce chaque sous-contrat sur un réseau d'agents. Les agents reçoivent et évaluent l'annonce. Les agents qui ont les ressources appropriées, l'expertise ou l'information requise envoient au gestionnaire des soumissions (bids) qui indiquent leurs capacités à réaliser la tâche annoncée. Le gestionnaire évalue les soumissions et accorde les tâches aux agents les mieux appropriés. Ces agents sont appelés des contractants (contractors). Enfin, gestionnaires et contractants échangent les informations nécessaires durant l'accomplissement des tâches. Par exemple, **Parunak** [173] a utilisé ce protocole pour développer un système de contrôle de production.

✓ Un autre important protocole de négociation a été proposé par Cammarata et ses collègues [174], [175] qui ont étudié les stratégies de coopération pour résoudre des conflits entre des plans d'un ensemble d'agents. Ces stratégies ont été appliquées au domaine du contrôle de trafic aérien avec le but de permettre à chaque agent (représentant un avion) de construire un plan de vol qui permettrait de garder une distance sécuritaire par rapport aux autres avions et de satisfaire des contraintes telles que "atteindre la destination désirée avec une consommation de carburant minimale". La stratégie choisie, appelée "centralisation de tâche" permettait aux agents impliqués dans une situation conflictuelle potentielle (des avions se rapprochant trop compte tenu de leurs caps respectifs) de choisir l'un d'eux pour résoudre le conflit. Cet agent agissait comme un planificateur centralisé et développait un plan multi-agent qui spécifiait les actions concurrentes de tous les avions impliqués. Les agents utilisaient la négociation pour déterminer qui était le plus apte à réaliser le plan. Cette aptitude était évaluée à partir de divers critères permettant d'identifier par exemple l'agent le mieux informé ou celui qui était le plus contraint. Les protocoles de négociation précédents supposent que les agents sont coopératifs, et donc qu'ils poursuivent un but commun.

- ✓ Le protocole multi-stage [106], [176] développé pour résoudre de façon coopérative des conflits dans l'allocation de ressources. Le domaine d'application était celui de la surveillance et du contrôle dans un système de communication complexe. Le protocole débute par la génération d'un plan initial et consiste en plusieurs cycles permettant d'envoyer des requêtes pour des buts secondaires, l'analyse locale, la génération de plans alternatifs et l'envoi de réponses.

## 6. Conclusion

A travers ce chapitre, nous avons présenté une synthèse de l'état de l'art dans le domaine de planification qui depuis sont apparition, est passé par plusieurs versions parmi lesquelles on trouve la planification distribuée qui est une extension de la planification dans le cadre des systèmes multi-agent. Cette extension a enrichi le domaine de la planification suite à l'exploitation des avantages du paradigme multi-agent. Cependant, l'utilisation de ce dernier a introduit de nouveaux problèmes inexistants dans la version classique de la planification.

Dans le chapitre suivant nous proposerons une approche qui traite l'un de ces problèmes.

# Chapitre 3
# L'approche proposée

## 1. Introduction

Le paradigme agent revêt de plus en plus d'importance pour sa capacité à aborder les systèmes complexes caractérisés par l'indéterminisme, l'émergence et l'évolution imprédictible. Il est très efficace pour gérer la nature hétérogène des composantes d'un système, pour modéliser les interactions entre les composantes de ce dernier et pour tenter de comprendre les phénomènes émergents qui en découlent. Ceci est lié au fait que l'agent possède un comportement, caractérisé principalement par quatre propriétés [5]:

- Autonomie ou proactivité : capacité à agir sans intervention extérieure, prise d'initiative.
- Sensibilité : capacité à percevoir l'environnement ou les autres agents.
- Localité : limitation de la perception et des actions.
- Flexibilité : réaction aux changements perçus.

En effet, l'agent ne se limite pas seulement à réagir aux invocations de méthodes spécifiques, comme il est souvent le cas dans le paradigme objet, mais également à tout autre changement observable dans son environnement. La prise en compte de ces changements se traduit automatiquement par un ensemble d'actions nouvelles que l'agent doit exécuter. La détermination de ces actions dépend de la nature de l'agent [6]. En effet si, par exemple, l'agent est rationnel, les actions à déterminer ne doivent pas être en opposition avec la fonction d'utilité de l'agent, si l'agent est avec but, ces actions ne doivent pas être en opposition avec le but de l'agent, si l'agent est réactif avec modèle, ces actions sont prédéterminée par un ensemble de règles, etc.

Le comportement de l'agent est ainsi source d'avantages mais les actions nouvelles à exécuter par l'agent, afin de prendre en considération les changements imprédictibles qui caractérisent son environnement, peuvent créer un problème lors de la planification distribuée. En effet dans la planification classique, l'ensemble des actions à planifier est défini auparavant et ne subit aucun changement assurant ainsi, une fiabilité du plan généré jusqu'à la fin de son exécution. Alors que dans la planification distribuée, chaque agent peut avoir des changements dans son ensemble d'actions à planifier, suite aux changements imprédictibles de son environnement.

A cause des changements survenus sur l'ensemble des actions, le plan que l'agent était entrain d'exécuter devient obsolète car il ne prend pas en considération les nouvelles actions à exécuter par l'agent, afin de prendre en considération les changements imprédictible de son environnement. L'agent se trouve par conséquent, contraint de générer un nouveau plan. De ce fait, la réflexion vers une approche de planification dynamique permet de générer, à tout moment et au fur et à mesure des changements, de nouveaux plans pour prendre en considération les nouvelles actions s'impose d'elle-même. Ceci représente le point focal de notre travail, dans lequel, nous proposons une nouvelle approche de planification dynamique distribuée capable de prendre en considération les changements pouvant survenir sur l'ensemble des actions à planifier. Notre approche s'intègre dans le contexte de la planification distribuée pour des plans distribués où chaque agent peut produire ses propres plans. Selon notre approche la génération des plans est basée sur la satisfaction des contraintes par l'utilisation des algorithmes génétiques.

2. **L'approche proposée**

Notre approche consiste à générer, d'une façon dynamique, les plans par chaque agent, dans le but de prendre en considération les changements imprédictibles de son environnement. Selon notre approche chaque agent doit générer le plan le plus adéquat, dans lequel il établit l'ordre, de l'ensemble des actions qu'il doit exécuter à un instant donné afin de répondre de la manière la plus satisfaisante à un ensemble de contraintes données imposées par le système, par la coordination entre les agents ou par lui-même. Pour cela on utilise les algorithmes génétiques ou la fonction de fitness proposée est définie par l'ensemble de contraintes que l'agent doit satisfaire dans le plan qu'il doit générer.

Au départ à l'instant $t_0$, chaque agent génère un ensemble de plans (population initiale) générés de manière aléatoire où chaque plan est une suite d'actions à exécuter dans un ordre différent des autres plans. A partir de la population initiale, chaque agent, à travers l'utilisation des algorithmes génétiques, génère des sous populations avec lesquelles il se rapproche, au fur et à mesure de la satisfaction de la fonction de fitness. Ceci se répète jusqu'à l'obtention du meilleur plan à exécuter. Pendant l'exécution de ce dernier, si l'agent se retrouve devant un changement, vis-à-vis duquel, de nouvelles actions doivent être exécutées, un autre plan, sera généré d'une façon dynamique pour prendre en considération ces actions. Dans ce nouveau plan, l'agent doit établir le meilleur ordre entre les anciennes actions non exécutées de l'ancien plan et les nouvelles actions engendrées par les changements de façon à satisfaire l'ensemble des contraintes.

De ce fait, la génération des plans se répète d'une façon récursive, prenant, à chaque fois, comme nouvel état initial; l'état dans lequel l'ensemble des actions de l'agent subit un changement.

Selon notre approche, le problème de planification à l'instant t est défini par un ensemble d'agents, l'état courant du système et le mécanisme de coordination à adopter en cas de conflit entre les agents lors de l'exécution de leurs plan. Chaque agent est défini par :

- L'ensemble d'action à exécuter à l'instant t. Chaque action est caractérisée par un état qui indique que cette action est réalisé ou non à l'instant t
- L'ensemble de contraintes à satisfaire pendant la génération du plan à l'instant t. Chaque contrainte est définie par un ensemble de variables, par une fonction de fitness et par un coefficient qui indique le poids de la contrainte
- Une fonction de fitness que l'agent doit optimiser lors de la génération de son plan. Elle est définie sur la base des fonctions de fitness de contraintes
- L'état courant de l'agent
- Une fonction de révision des actions qui permet la mise à jour des actions de l'agent à tout moment

### 2.1. Définition formel du problème

Le problème de la planification selon notre approche peut être défini comme suit :

$\Pi_t$ = (N, $S_{st}$, $S_{sf}$, Coordination) avec :

- N représente un ensemble d'agent N= {$A_1$, $A_2$………..,$A_n$} où chaque agent $A_i$ est un agent défini comme suit: $A_i$ = {$\alpha_{it}$, $C_{it}$, f\_$A_i$, $S_{it}$, $S_{if}$, Rev\_$\alpha_i$} avec :
  - $\alpha_{it}$ représente l'ensemble des actions de l'agent $A_i$ à l'instant t. $\alpha_{it}$ = {($\alpha_{ij}$, e)} avec :
    - $\alpha_{ij}$ représente l'action j de l'agent i
    - e représente l'état qui indique que cette action est réalisé ou non à l'instant t.
  - $C_{it}$ représente l'ensemble des contraintes à satisfaire par l'agent $A_i$. $C_{it}$ = {(Var\_$C_{ij}$, f\_$C_{ij}$, Coef\_$C_{ij}$)} avec :
    - $C_{ij}$ représente la contrainte j de l'agent i
    - Var\_$C_{ij}$ représente l'ensemble de variables sur le quel la contrainte $C_{ij}$ est défini
    - f\_$C_{ij}$ représente la fonction de fitness correspondante à la contrainte $C_{ij}$
    - Coef\_$C_{ij}$ représente le coefficient de la contrainte $C_{ij}$, il indique l'importance de cette dernière
  - f\_$A_i$ représente la fonction de fitness que l'agent $A_i$ devra optimiser.
    $$f\_A_i = \sum_{j=1}^{n}(f\_Cij * Coef\_Cij) / \sum_{j=1}^{n} Coef\_Cij$$
  - Rev\_$\alpha_i$ : représente la fonction de révision des actions qui peut changer les actions de l'agent $A_i$ en fonction de sa nature
  - $S_{ti}$ représente l'état courant de l'agent i, cet état change, au fur et à mesure, avec l'exécution des actions
  - $S_{fi}$ représente l'état final de l'agent i
- $S_{ts}$ représente l'état courant du système. Cet état change au fur et à mesure, avec l'exécution des actions par l'ensemble d'agent N.
- $S_{fs}$ représente l'état final du système.
- Coordination : représente le processus de coordination choisi en cas de conflit entre les agents pendant l'exécution de leurs plans

La solution du problème à l'instant t est définie comme suit :

- $P_t$ = {$P_{1t}$, $P_{2t}$,……….,$P_{nt}$} avec :
- $P_{it}$ représente le meilleur plan de l'agent $A_i$ pour exécuter ses actions $\alpha_{it}$ à l'instant t.

### 2.2. Génération des plans initiaux $P_0$

Chaque agent $A_i$ commence par la génération d'un plan initial $P_{i0}$ dans lequel il doit établir le meilleur ordre entre son ensemble initial d'actions $\alpha_{i0}$ de façon à satisfaire l'ensemble des contraintes. Pour cela les algorithmes génétiques sont utilisés, où la fonction de fitness proposée est définie par l'ensemble de contraintes que l'agent doit satisfaire dans le plan qu'il doit générer.

L'agent $A_i$, dans un premier temps, génère une population initiale $Pop_{i0}$, dans laquelle, il construit un nombre important d'individus. Chaque individu correspond à un plan $P_{ij}$ (le plan j de l'agent i) qui contient toutes les actions de l'ensemble initial d'actions $\alpha_{i0}$ dans un ordre différent des autres plans. La génération de la population initiale se fait d'une façon aléatoire. Dans le but de trouver une population plus performante $Pop_{i1}$ qui satisfera mieux l'ensemble des contraintes l'agent $A_i$ applique les étapes suivantes sur la $Pop_{i0}$:

- Tout d'abord, chaque plan $P_{ij}$ de la population $Pop_{i0}$ est évalué. Pour cela, la valeur de la fonction de fitness $f\_A_i$ est calculée.
- Puis, une étape de sélection est appliquée. Cette étape permet d'éliminer les moins bons plans de la $Pop_{i0}$ et de garder uniquement les meilleurs en fonction de leur évaluation.
- L'étape suivante consiste à croiser les plans précédemment sélectionnés pour obtenir la nouvelle population $Pop_{i1}$. Deux plans (parents) sont donc choisis pour appliquer un opérateur de croisement afin d'obtenir un nouveau plan (descendant). Il existe de nombreuses techniques de croisement, dans notre approche, nous utiliserons le « *crossover* en un point ». Cet opérateur consiste à recopier une partie du parent 1 et une partie du parent 2 pour obtenir un nouvel individu. Le point de séparation des parents est appelé point de croisement. Il faut cependant faire attention à ne pas répéter la même action (on ne recopie pas les actions déjà mise dans le plan), et à ne pas oublier des actions (on rajoute à la fin les actions non prises en compte).
- Pour diversifier les solutions au fur et à mesure des générations une étape de mutation est utilisée. Cette mutation consiste à modifier aléatoirement une petite partie d'un caractère dans certains individus de la nouvelle génération. Cette étape est effectuée avec une très faible probabilité, et consiste par exemple à échanger deux actions consécutives dans un plan.

A partir de la population $Pop_{i1}$ obtenue, les étapes décrites plus haut seront réappliquées, dans la limite du nombre de générations possibles « k » précisé par le programme. Ceci, dans le but d'obtenir une population dans laquelle se trouve un plan qui satisfasse totalement l'ensemble des contraintes. Ce plan sera considéré comme plan initial $p_{i0}$ que l'agent $A_i$ doit exécuter. Dans le cas où ce plan n'est pas obtenu dans la limite des générations possibles, le meilleur plan obtenu dans la dernière génération (Pop$_{ik}$) sera pris en tant que plan initial.

### 2.3. Génération dynamique des plans

Chaque agent A$_i$ commence par l'exécution de son plan initial P$_{i0}$. Si pendant l'exécution l'agent se retrouve, à un instant donné t, devant un changement, vis-à-vis duquel, de nouvelles actions doivent être exécutées, un autre plan P$_{it}$, sera généré par l'agent d'une façon dynamique pour prendre en considération ces nouvelles actions. Pour la génération de ce nouveau plan P$_{it}$, l'agent applique les mêmes étapes appliquées pour la génération du plan initial P$_{i0}$ sachant que :

- l'ensemble d'actions à planifier α$_{it}$ sera égal à l'ensemble des anciennes actions non exécutées de l'ancien plan plus les nouvelles actions engendrées par les changements. Dans le nouveau plan P$_{it}$ l'agent doit établir le meilleur ordre entre les actions de l'ensemble α$_{it}$ de façon à satisfaire l'ensemble des contraintes.
- la population initiale Pop$_{i0}$ sera générée sur la base de ce nouvel ensemble d'action α$_{it}$
- l'état initial de l'agent pour ce plan sera l'état de l'agent à l'instant t S$_{it}$
- l'état initial de système pour ce plan sera l'état de système à l'instant t S$_{st}$

La figure suivante (figure 6) représente le processus de notre approche

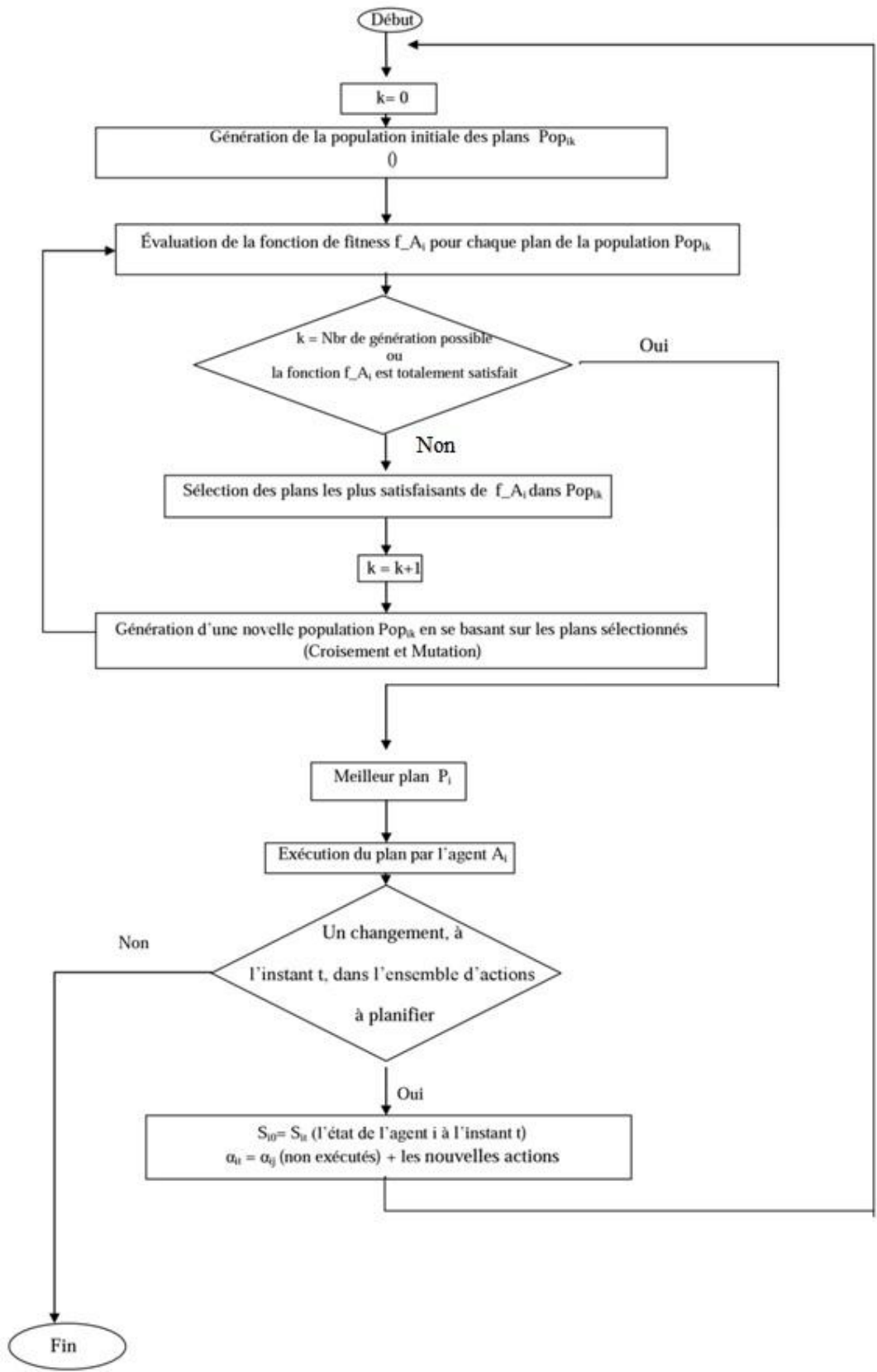

Figure 6 : Le processus de l'approche proposée.

### 3. Conclusion

Dans ce chapitre, nous avons proposé une nouvelle approche de planification distribuée dynamique, capable de prendre en considération les changements que l'agent introduit sur son ensemble des actions à planifier dans le but de prendre en compte les changements qui surviennent dans son environnement.

Notre approche consiste à faire générer, un nouveau plan par chaque agent, à chaque fois qu'il y' a un changement dans son ensemble d'actions à planifier. Ceci dans le but de prendre en considération les nouvelles actions introduites dans son nouveau plan. Dans ce nouveau plan l'agent prend, à chaque fois, comme nouvel ensemble d'actions à planifier l'ensemble des anciennes actions non exécutées de l'ancien plan et les nouvelles actions engendrées par les changements et comme nouvel état initial; l'état dans lequel l'ensemble des actions de l'agent subit un changement

Dans le but de valider notre approche, nous l'appliquons sur une étude de cas concrète dans le chapitre suivant.

.

# Chapitre 4
# Etude de cas

**1. Introduction**

Pour valider notre approche, nous l'appliquons sur une étude de cas concrète: DPDP (Dynamic Pick and Delivery Problem). L'objectif de ce système est de distribuer des articles $Art_j$ stockés dans des points de distribution $S_j$ à un ensemble de clients positionnés dans des points $T_k$. Le système reçoit à chaque fois des requêtes que les agents doivent réaliser à travers l'exécution d'un ensemble d'actions. Chaque requête est définie par le client $T_k$ demandeur, l'article $Art_j$ demandé, le point dans lequel l'article est stocké $S_j$, la quantité de l'article demandé ainsi que l'agent $A_i$ concerné par cette requête.

Selon notre approche, chaque agent génère le meilleur plan dans lequel il établit un ordre pour l'ensemble de ses actions nécessaires pour accomplir ses requêtes afin répondre, de la façon la plus satisfaisante, à un ensemble de contraintes (minimiser la distance, réduire le nombre d'obstacles lors de la distribution). Pendant l'exécution du plan généré, si l'agent se retrouve devant un changement, vis-à-vis duquel, de nouvelles actions doivent être exécutées, un autre plan, sera généré d'une façon dynamique pour prendre en considération ces actions. Dans ce nouveau plan, l'agent doit établir le meilleur ordre entre les anciennes actions non exécutées de l'ancien plan et les nouvelles actions engendrées par les changements de façon à satisfaire l'ensemble des contraintes.

De ce fait, la génération des plans se répète d'une façon récursive, prenant comme un nouvel état initial l'état dans lequel l'ensemble d'actions de l'agent change.

L'état de système est défini à chaque fois par la position des clients $T_i$, et la position des points de distribution $S_i$ et par les requêtes à accomplir. L'état de chaque agent est défini à chaque fois par la position de l'agent, le niveau de la batterie et le taux de ses actions réalisées.

Le choix de l'étude de cas a été fondé sur le fait qu'elle permet l'illustration de différentes étapes de l'approche proposée d'une manière simple et claire. Dans ce qui suit, nous allons dans un premier temps $t_0$ appliquer l'approche sur un état de système donné pour lequel les agents génèrent les plans initiaux. Dans un deuxième temps, nous proposerons un changement qui sera appliqué pendant l'exécution des plans générés dans le but de montrer le dynamisme de l'approche

Dans cette étude de cas nous avons adopté :

- Trois agents réactifs avec modèle dont la détermination des actions est prédéfinie par un ensemble de règles Reg:
    - Reg1 = Pour chaque requête $R_m$= (($S_j$, $Art_j$, $T_k$), quantité, $A_i$, False) l'agent $A_i$ doit exécuté les actions suivantes : (Move ($S_j$), False), (Take ($S_j$, $Art_j$, quantité), False), (Move ($T_k$), False), (Delivery ($T_k$, $Art_j$, quantité), False).
    - Reg2 = Lorsque la batterie d'un agent $A_i$ est égale au 1/10 de sa charge maximale, l'agent doit exécuté l'action suivante : Charg_batterie ((Position Chargeur), False).
- Le processus de coordination choisi en, cas de conflit, est le processus de coordination par arbitrage où l'agent arbitre ou médiateur dispose des différents points de conflit possibles et la manière de résoudre ces conflits
- Pour l'algorithme génétique, nous avons adopté un codage réel et nous avons choisi les valeurs suivantes : Taille de la population initiale = 20; nombre maximum de générations = 30; probabilité de mutation = 2%, pourcentage de sélection = 80%
- Pour l'implémentation nous avons utilisé la plate-forme JADE et pour la génération des plans nous avons utilisé la structure des listes.

2. **Le problème de planification à l'instant $t_0$**

Supposons à l'instant $t_0$ que nous ayons 7 clients $T_1, T_2…..T_7$ et 4 points de distributions $S_1, S_2…S_4$ (figure 7)

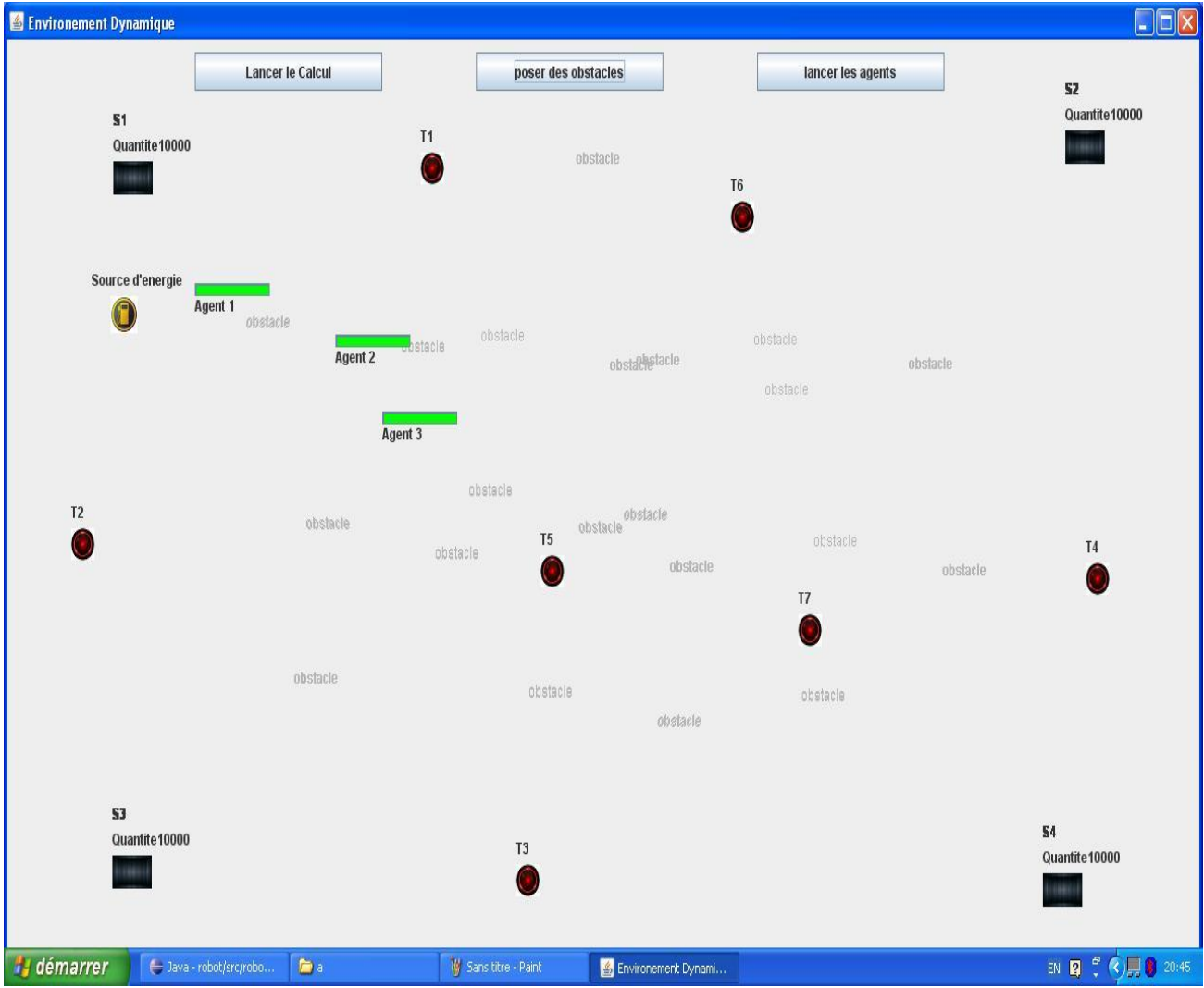

Figure 7 : L'état du système à l'instant $t_0$.

Le problème de planification à l'instant $t_0$ est défini comme suit:

$\Pi_0 = (N, S_{ts}, S_{fs}, Coordination)$ avec :

- $N = \{A_1, A_2, A_3\}$ avec :
    - $A_1 = \{\alpha_{10}, C_{10}, f\_A_1, S_{10}, S_{f1}, Rev\_\alpha_1\}$ avec:
        - $\alpha_{10} = \{(Move (S_1), False), (Take (S_1, Art_1, 100), False), (Move (T_3), False), (Delivery(T_3, Art_1, 100),False), (Move (S_3), False), (Take (S_3, Art_3, 50), False), (Move (T_4), False), (Delivery(T_4, Art_3, 50),False),$

- (Move ($S_2$), False), (Take ($S_2$, $Art_2$, 150), False), (Move ($T_1$), False), (Delivery($T_1$, $Art_2$, 150),False) }
    - $C_{10}$= {(Var_$C_{11}$=Dis (Point i, Point j),
      f_$C_{11}$=1/$\sum_{\substack{j=1 \\ i=1}}^{n}$ Dis(Point i, Point j) , Coef_$C_{11}$= 10),
      (Var_$C_{12}$=Nbr_Obs(Point i, Point j),
      f_$C_{12}$= 1/$\sum_{\substack{j=1 \\ i=1}}^{n}$ Nbr_Obs(Point i, Point j) , Coef_$C_{12}$=8)} / Point i,
      Point j peuvent êtres soit des points de stocke (S) où de requête (T).
    - f_$A_1$=( f_$C_{11}$ * Coef_$C_{11}$) + (f_$C_{12}$ * Coef_$C_{12}$)/ (Coef_$C_{11}$+ Coef_$C_{12}$)
    - $S_{10}$= {Position (200, 200), Batterie (100%), Taux_Actions_réalisées = 0}
    - $S_{f1}$= { Actions_non_réalisées = $\phi$ }
    - Rev_$\alpha_1$: Reg x P x $\alpha_1$ ---> $\alpha_{1j}$ où Reg représente l'ensemble de règles de l'agent $A_1$, P représente l'ensemble des perceptions de l'agent $A_1$, $\alpha_1$ représente l'ensemble d'actions de l'agent $A_1$
- $A_2$ = { $\alpha_{20}$, $C_{20}$, f_$A_2$, $S_{20}$, $S_{f2}$, Rev_$\alpha_2$} avec:
    - $\alpha_{20}$ = {(Move ($S_3$), False), ( Take ($S_3$, $Art_3$, 150), False), (Move ($T_2$), False), (Delivery($T_2$, $Art_3$, 150),False), (Move ($S_4$), False), (Take ($S_4$, $Art_4$, 50), False), (Move ($T_5$), False), (Delivery($T_5$, $Art_4$, 50),False), (Move ($S_2$), False), (Take ($S_2$, $Art_2$, 300), False), (Move ($T_7$), False), (Delivery($T_7$, $Art_2$, 300),False), (Move ($S_1$), False), (Take ($S_1$, $Art_1$, 200), False), (Move ($T_6$), False), (Delivery($T_6$, Art1, 200),False) }
    - $C_{20}$ = {(Var_$C_{21}$=Dis (Point i, Point j),
      f_$C_{21}$=1/$\sum_{\substack{j=1 \\ i=1}}^{n}$ Dis(Point i, Point j) , Coef_$C_{21}$= 5),
      (Var_$C_{22}$=Nbr_Obs(Point i, Point j),
      f_$C_{22}$=1/$\sum_{\substack{j=1 \\ i=1}}^{n}$ Nbr_Obs(Point i, Point j) , Coef_$C_{12}$=15)} / Point i,
      Point j peuvent êtres soit des points de stocke (S) où de requête (T).
    - f_$A_2$=( f_$C_{21}$ * Coef_$C_{21}$) + (f_$C_{22}$ * Coef_$C_{22}$)/ (Coef_$C_{21}$+ Coef_$C_{22}$)
    - $S_{20}$ = {Position (350, 240), Batterie (100%), Taux_Actions_réalisées = 0}
    - $S_{f2}$ = { Actions_non_réalisées = $\phi$ }

- Rev_$\alpha_2$: Reg x P x $\alpha_2$ ---> $\alpha_{2j}$ où Reg représente l'ensemble de règles de l'agent $A_2$, P représente l'ensemble des perceptions de l'agent $A_2$, $\alpha_2$ représente l'ensemble d'actions de l'agent $A_2$
- $A_3$ = { $\alpha_{30}$, $C_{30}$, f_$A_3$, $S_{30}$, $S_{f3}$, Rev_$\alpha_3$} avec:
  - $\alpha_{30}$ = {(Move ($S_2$), False), (Take ($S_2$, $Art_2$, 150), False), (Move ($T_3$), False), (Delivery($T_3$, $Art_2$, 150), False), (Move ($S_1$), False), (Take ($S_1$, $Art_1$, 150), False), (Move ($T_2$), False), (Delivery($T_2$, $Art_1$, 150), False), (Move ($S_4$), False), (Take ($S_4$, $Art_4$, 200), False), (Move ($T_1$), False), (Delivery($T_1$, $Art_4$, 200), False) }
  - $C_{30}$= {(Var_$C_{31}$=Dis (Point i, Point j), f_$C_{31}$=1/$\sum_{\substack{j=1 \\ i=1}}^{n}$ Dis(Point i , Point j ) , Coef_$C_{31}$= 9),
    (Var_$C_{32}$=Nbr_Obs(Point i, Point j), f_$C_{32}$=1/$\sum_{\substack{j=1 \\ i=1}}^{n}$ Nbr_Obs(Point i , Point j ) , Coef_$C_{32}$=3)} / Point i , Point j peuvent êtres soit des points de stocke (S) où de requête (T)
  - f_$A_3$= ( f_$C_{31}$ * Coef_$C_{31}$) + (f_$C_{32}$ * Coef_$C_{32}$)/ (Coef_$C_{31}$+ Coef_$C_{32}$)
  - $S_{30}$= {Position (400, 300), Batterie (100%), Taux_Actions_réalisées = 0}
  - $S_{f3}$= {Actions_non_réalisées = $\phi$}
  - Rev_$\alpha_3$: Reg x P x $\alpha_3$ ---> $\alpha_{3j}$ où Reg représente l'ensemble de règles de l'agent $A_3$, P représente l'ensemble des perceptions de l'agent $A_3$, $\alpha_3$ représente l'ensemble d'actions de l'agent $A_3$
  - 

- $S_{0s}$ = {S= [$S_1$(P=(200, 150), Q=10000), $S_2$(P=(1800, 120), Q=10000), $S_3$(P=(180, 1000), Q=10000), $S_4$ (P(1750, 1010), Q=10000)], T= [$T_1$(P=(500,100)), $T_2$(P=(100,800)), $T_3$(P=(800, 800)), $T_4$(P=(1800, 720)), $T_5$(P=(650, 670)), $T_6$(P=(800, 320)) $T_7$(P=(850, 700))], R= [$R_1$= (($S_1$, $Art_1$, $T_3$), 100, $A_1$, False), $R_2$=(($S_3$, $Art_3$, $T_4$), 50, $A_1$, False), $R_3$=(($S_2$, $Art_2$, T1), 150, $A_1$, False), $R_4$= (($S_3$, $Art_3$, $T_2$), 150, $A_2$, False), $R_5$=((S4, $Art_4$ ,T5), 50, $A_2$, False), $R_6$=(($S_2$, $Art_2$, $T_7$), 300, $A_2$, False), $R_7$=(($S_1$, $Art_1$, $T_6$), 200, $A_2$, False), $R_8$= (($S_2$, $Art_2$,$T_3$), 150, $A_3$, False), $R_9$=(($S_1$, $Art_1$, $T_2$), 150, $A_3$, False), $R_{10}$=(($S_4$, $Art_4$, $T_1$), 200, $A_3$, False)]}
- $S_{fs}$ = {Requêtes_non_réalisées = $\phi$}
- Coordination = Coordination par arbitrage

## 3. Application de notre approche
### 3.1. Génération d'un plan initial par chaque agent

Pour générer son plan initial, chaque agent $A_i$, commence par la génération de la population initiale $Pop_{i0}$ à partir de son ensemble d'actions initiales $α_{i0}$. Les figures 8, 9, 10 représentent dix plans de la population initiale des agents $A_1$, $A_2$, $A_3$ respectivement.

```
Plan 0
(MoveS1,false)(TakeS1,Article1,100,false)(MoveS2,false)(TakeS2,Article2,150,false)
(Move T3,false)(DelevryT3,Article 1,100,false)(Move S3,false)
(TakeS3,Article 3,50,false) (Move T4,false)(DelevryT4,Article 3,50,false)
(Move T1,false)(DelevryT1,Article 2,150,false)

Plan 1
(Move S2,false)(TakeS2,Article 2,150,false)(Move S3,false)(TakeS3,Article
3,50,false)(Move S1,false)(TakeS1,Article 1,100,false)(Move T3,false) (DelevryT3,Article
1,100,false)(Move T4,false)(DelevryT4,Article 3,50,false)
(Move T1,false)(DelevryT1,Article 2,150,false)

Plan 2
(Move S2,false)(TakeS2,Article 2,150,false)(Move S1,false)
(TakeS1,Article 1,100,false)(Move S3,false)(TakeS3,Article 3,50,false)
(Move T3, false)  (DelevryT3,Article 1,100,false)(Move T1,false)
(DelevryT1,Article 2,150,false)(Move T4,false)(DelevryT4,Article 3,50,false)

Plan 3
(Move S2,false)(TakeS2,Article 2,150,false)(Move T1,false)
(DelevryT1,Article 2,150,false)(Move S3,false)(TakeS3,Article 3,50,false)
(Move S1,false)(TakeS1,Article 1,100,false)(Move T4,false)
(DelevryT4,Article 3,50,false)(Move T3,false)(DelevryT3,Article 1,100,false)

Plan 4
(Move S2,false)(TakeS2,Article 2,150,false)(Move T1,false)(DelevryT1,Article
2,150,false)(Move S3,false)
(TakeS3,Article 3,50,false)(Move S1,false)(TakeS1,Article 1,100,false)(Move T3,false)
(DelevryT3,Article 1,100,false)  (Move T4,false)(DelevryT4,Article 3,50,false)

Plan 5
(Move S3,false)(TakeS3,Article 3,50,false)(Move S1,false)
(TakeS1,Article 1,100,false)  (Move T3,false)(DelevryT3,Article 1,100,false)
(Move T4,false)(DelevryT4,Article 3,50,false)(Move S2,false)
(TakeS2,Article 2,150,false)(Move T1,false)(DelevryT1,Article 2,150,false)

Plan 6
(Move S2,false)(TakeS2,Article 2,150,false)(Move S3,false)
(TakeS3,Article 3,50,false)(Move T4,false)(DelevryT4,Article 3,50,false)
(Move S1,false)(TakeS1,Article 1,100,false)(Move T3,false)
(DelevryT3,Article 1,100,false)(Move T1,false)(DelevryT1,Article 2,150,false)

Plan 7
(Move S1,false)(TakeS1,Article 1,100,false)(Move S3,false)
(TakeS3,Article 3,50,false)(Move S2,false)(TakeS2,Article 2,150,false)
(Move T4,false)(DelevryT4,Article 3,50,false)(Move T3,false)
(DelevryT3,Article 1,100,false)(Move T1,false)(DelevryT1,Article 2,150,false)

Plan 8
(Move S3,false)(TakeS3,Article 3,50,false)(Move T4,false)
(DelevryT4,Article 3,50,false)(Move S1,false)(TakeS1,Article 1,100,false)
(Move T3,false)(DelevryT3,Article 1,100,false)(Move S2,false)
(TakeS2,Article 2,150,false)(Move T1,false)(DelevryT1,Article 2,150,false)

Plan 9
(Move S2,false)(TakeS2,Article 2,150,false)(Move S3,false)
(TakeS3,Article 3,50,false)(Move S1,false)(TakeS1,Article 1,100,false)
(Move T3,false)(DelevryT3,Article 1,100,false)(Move T1,false)
(DelevryT1,Article 2,150,false)(Move T4,false)(DelevryT4,Article 3,50,false)
```

Figure 8: Dix plans de la population initiale de l'agent $A_1$.

```
Plan 0
(Move S1,false)(TakeS1,Article 1,200,false)(Move T6,false)
(DelevryT6,Article 1,200,false) (Move S3,false)(TakeS3,Article 3,150,false)
(Move S4,false)(TakeS4,Article 4,50,false)(Move T2,false)
(DelevryT2,Article 3,150,false)(Move S2,false)(TakeS2,Article 2,300,false)
(Move T5,false)(DelevryT5,Article 4,50,false)(Move T7,false)
(DelevryT7,Article 2,300,false)

Plan 1
(Move S3,false)(TakeS3,Article 3,150,false)(Move S2,false)
(TakeS2,Article 2,300,false)(Move S1,false)(TakeS1,Article 1,200,false)
(Move S4,false)(TakeS4,Article 4,50,false)(Move T2,false)
(DelevryT2,Article 3,150,false)(Move T5,false)(DelevryT5,Article 4,50,false)
(Move T6,false)(DelevryT6,Article 1,200,false)(Move T7,false)
(DelevryT7,Article 2,300,false)

Plan 2
(Move S2,false)(TakeS2,Article 2,300,false)(Move S3,false)(TakeS3,Article
3,150,false)(Move T7,false)(DelevryT7,Article 2,300,false)(Move
T2,false)(DelevryT2,Article 3,150,false)(Move S4,false)(TakeS4,Article 4,50,false)(Move
S1,false)(TakeS1,Article 1,200,false)(Move T6,false)(DelevryT6,Article 1,200,false)(Move
T5,false)(DelevryT5,Article 4,50,false)

Plan 3
(Move S3,false)(TakeS3,Article 3,150,false)(Move S1,false)
(TakeS1,Article 1,200,false)(Move T6,false)(DelevryT6,Article 1,200,false)
(Move S4,false)(TakeS4,Article 4,50,false)(Move T5,false)
(DelevryT5,Article 4,50,false)(Move S2,false)(TakeS2,Article 2,300,false)
(Move T7,false)(DelevryT7,Article 2,300,false)(Move T2,false)
(DelevryT2,Article 3,150,false)

Plan 4
(Move S3,false)(TakeS3,Article 3,150,false)(Move S1,false)
(TakeS1,Article 1,200,false)(Move T6,false)(DelevryT6,Article 1,200,false)(Move
S4,false)(TakeS4,Article 4,50,false)(Move T5,false)(DelevryT5,Article 4,50,false)(Move
T2,false)(DelevryT2,Article 3,150,false)(Move S2,false)(TakeS2,Article 2,300,false)(Move
T7,false)(DelevryT7,Article 2,300,false)

Plan 5
(Move S3,false)(TakeS3,Article 3,150,false)(Move T2,false)(DelevryT2,Article
3,150,false)(Move S4,false)(TakeS4,Article 4,50,false)(Move S1,false)
(TakeS1,Article 1,200,false)(Move S2,false)(TakeS2,Article 2,300,false)
(Move T6,false)(DelevryT6,Article 1,200,false)(Move T7,false)
(DelevryT7,Article 2,300,false)(Move T5,false)(DelevryT5,Article 4,50,false)

Plan 6
(Move S3,false)(TakeS3,Article 3,150,false)(Move S2,false)(TakeS2,Article
2,300,false)(Move S1,false)(TakeS1,Article 1,200,false)
(Move S4,false)(TakeS4,Article 4,50,false)(Move T2,false)
(DelevryT2,Article 3,150,false)(Move T5,false)(DelevryT5,Article 4,50,false)
(Move T6,false)(DelevryT6,Article 1,200,false)(Move T7,false)
(DelevryT7,Article 2,300,false)

Plan 7
(Move S1,false)(TakeS1,Article 1,200,false)(Move S2,false)(TakeS2,Article
2,300,false)(Move S3,false)(TakeS3,Article 3,150,false)
(Move S4,false)(TakeS4,Article 4,50,false)(Move T7,false)
(DelevryT7,Article 2,300,false)(Move T6,false)(DelevryT6,Article 1,200,false)
(Move T2,false)(DelevryT2,Article 3,150,false)(Move T5,false)(DelevryT5,Article
4,50,false)

Plan 8
(Move S1,false)(TakeS1,Article 1,200,false)(Move S4,false)(TakeS4,Article
4,50,false)(Move S2,false)(TakeS2,Article 2,300,false)
(Move T7,false)(DelevryT7,Article 2,300,false)(Move T5,false)
(DelevryT5,Article 4,50,false)(Move S3,false)(TakeS3,Article 3,150,false)
(Move T6,false)(DelevryT6,Article 1,200,false)(Move T2,false)
(DelevryT2,Article 3,150,false)

Plan 9
(Move S1,false)(TakeS1,Article 1,200,false)(Move S2,false)
(TakeS2,Article 2,300,false)(Move S3,false)(TakeS3,Article 3,150,false)
(Move T2,false)(DelevryT2,Article 3,150,false)(Move S4,false)
(TakeS4,Article 4,50,false)(Move T5,false)(DelevryT5,Article 4,50,false)
(Move T6,false)(DelevryT6,Article 1,200,false)(Move T7,false)
(DelevryT7,Article 2,300,false)
```

Figure 9: Dix plans de la population initiale de l'agent A$_2$.

```
Plan 0
(Move S2,false)(TakeS2,Article 2,150,false)(Move S1,false)
(TakeS1,Article 1,150,false)(Move T2,false)(DelevryT2,Article 1,150,false)
(Move S4,false)(TakeS4,Article 4,200,false)(Move T1,false)
(DelevryT1,Article 4,200,false)(Move T3,false)(DelevryT3,Article 2,150,false)
Plan 1
(Move S2,false)(TakeS2,Article 2,150,false)(Move S1,false)
(TakeS1,Article 1,150,false)(Move S4,false)(TakeS4,Article 4,200,false)
(Move T2,false)(DelevryT2,Article 1,150,false)(Move T3,false)
(DelevryT3,Article 2,150,false)(Move T1,false)(DelevryT1,Article 4,200,false)
Plan 2
(Move S4,false)(TakeS4,Article 4,200,false)(Move S1,false)
(TakeS1,Article 1,150,false)(Move S2,false)(TakeS2,Article 2,150,false)
(Move T2,false)(DelevryT2,Article 1,150,false)(Move T3,false)
(DelevryT3,Article 2,150,false)(Move T1,false)(DelevryT1,Article 4,200,false)
Plan 3
(Move S1,false)(TakeS1,Article 1,150,false)(Move S2,false)
(TakeS2,Article 2,150,false)(Move S4,false)(TakeS4,Article 4,200,false)
(Move T2,false)(DelevryT2,Article 1,150,false)(Move T3,false)
(DelevryT3,Article 2,150,false)(Move T1,false)(DelevryT1,Article 4,200,false)
Plan 4
(Move S1,false)(TakeS1,Article 1,150,false)(Move S2,false)
(TakeS2,Article 2,150,false)(Move T3,false)(DelevryT3,Article 2,150,false)
(Move S4,false)(TakeS4,Article 4,200,false)(Move T2,false)
(DelevryT2,Article 1,150,false)(Move T1,false)(DelevryT1,Article 4,200,false)
Plan 5
(Move S2,false)(TakeS2,Article 2,150,false)(Move S4,false)
(TakeS4,Article 4,200,false)(Move S1,false)(TakeS1,Article 1,150,false)
(Move T1,false)(DelevryT1,Article 4,200,false)(Move T2,false)
(DelevryT2,Article 1,150,false)(Move T3,false)(DelevryT3,Article 2,150,false)
Plan 6
(Move S2,false)(TakeS2,Article 2,150,false)(Move T3,false)(DelevryT3,Article 2,150,false)(Move S1,false)(TakeS1,Article 1,150,false)
(Move S4,false)(TakeS4,Article 4,200,false)(Move T1,false)
(DelevryT1,Article 4,200,false)(Move T2,false)(DelevryT2,Article 1,150,false)
Plan 7
(Move S2,false)(TakeS2,Article 2,150,false)(Move S4,false)
(TakeS4,Article 4,200,false)(Move T1,false)(DelevryT1,Article 4,200,false)
(Move S1,false)(TakeS1,Article 1,150,false)(Move T3,false)
(DelevryT3,Article 2,150,false)(Move T2,false)(DelevryT2,Article 1,150,false)
Plan 8
(Move S4,false)(TakeS4,Article 4,200,false)(Move T1,false)
(DelevryT1,Article 4,200,false)(Move S2,false)(TakeS2,Article 2,150,false)
(Move T3,false)(DelevryT3,Article 2,150,false)(Move S1,false)
(TakeS1,Article 1,150,false)(Move T2,false)(DelevryT2,Article 1,150,false)
Plan 9
(Move S4,false)(TakeS4,Article 4,200,false)(Move T1,false)
(DelevryT1,Article 4,200,false)(Move S2,false)(TakeS2,Article 2,150,false)
(Move T3,false)(DelevryT3,Article 2,150,false)(Move S1,false)
(TakeS1,Article 1,150,false)(Move T2,false)(DelevryT2,Article 1,150,false)
```

Figure 10: Dix plans de la population initiale de l'agent $A_3$.

Dans le but de trouver une population plus performante $Pop_{i1}$ qui satisfera mieux l'ensemble des contraintes l'agent $A_i$ applique les étapes suivantes sur la $Pop_{i0}$:

- L'évaluation: cette étape consiste à calculer la fonction de fitness $f\_A_i$ de chaque plan $P_{ij}$ de la population initiale $Pop_{i0}$. Les figures 11, 12, 13 représentent l'évaluation de dix plans de la population initiale des agents $A_1$, $A_2$, $A_3$ respectivement.

```
Plan 0      F_C1 =3.6583253244223507E-4  F_C2 =0.01                  F_A =4.531462743252149E-5
Plan 1      F_C1 =6.323401357583406E-4   F_C2 =0.01                  F_A =7.781241907869489E-5
Plan 2      F_C1 =5.897754111634356E-4   F_C2 =0.01                  F_A =7.265073595371868E-5
Plan 3      F_C1 =6.323401357583406E-4   F_C2 =0.01                  F_A =7.781241907869489E-5
Plan 4      F_C1 =5.77533185332251E-4    F_C2 =0.027777777777777776  F_A =7.181835083820729E-5
Plan 5      F_C1=3.9932094414700796E-4   F_C2 =0.011363636363636364  F_A =4.9480429438769634E-5
Plan 6      F_C1 =4.143145586936494E-4   F_C2 =0.01                  F_A =5.1258392381236724E-5
Plan 7      F_C1 =4.143145586936494E-4   F_C2 =0.01                  F_A =5.1258392381236724E-5
Plan 8      F_C1 =3.7730500878527685E-4  F_C2 =0.00980392156862745   F_A =4.6713680801655446E-5
Plan 9      F_C1 =3.689321454479055E-4   F_C2 =0.011111111111111112  F_A =4.573685774725651E-5
```

Figure 11 : L'évaluation de dix plans de la population initiale de l'agent $A_1$.

```
Plan 0      F_C1 =2.1472768420252827E-4  F_C2 =0.00641025641025641   F_A =2.6618050860961023E-5
Plan 1      F_C1 =2.6595279043151345E-4  F_C2 =0.00641025641025641   F_A =3.290282547228916E-5
Plan 2      F_C1 =3.116105495296216E-4   F_C2 =0.00641025641025641   F_A =3.8483634339865544E-5
Plan 3      F_C1 =2.705071325893138E-4   F_C2 =0.00641025641025641   F_A =3.346039186200589E-5
Plan 4      F_C1 =3.8486730545899116E-4  F_C2 =0.006756756756756757  F_A =4.743296349885949E-5
Plan 5      F_C1 =3.385426476167838E-4   F_C2 =0.006535947712418301  F_A =4.177685179203307E-5
Plan 6      F_C1 =2.6005714273620996E-4  F_C2 =0.008264462809917356  F_A =3.2253413988792484E-5
Plan 7      F_C1 =2.2290002356262293E-4  F_C2 =0.006024096385542169  F_A =2.7607127328053214E-5
Plan 8      F_C1 =4.1487733971837395E-4  F_C2 =0.011111111111111112  F_A =5.138004806751753E-5
Plan 9      F_C1 =3.0030781367048617E-4  F_C2 =0.008333333333333333  F_A =3.7203303427402686E-5
```

Figure 12 : L'évaluation de dix plans de la population initiale de l'agent $A_2$.

```
Plan 0      F_C1 =6.186662433765832E-4   F_C2 =0.020833333333333332  F_A =7.676338940901854E-5
Plan 1      F_C1 =5.678571076814047E-4   F_C2 =0.020833333333333332  F_A =7.050171962903428E-5
Plan 2      F_C1 =6.662723350575081E-4   F_C2 =0.020833333333333332  F_A =8.262344529228388E-5
Plan 3      F_C1 =6.186662433765832E-4   F_C2 =0.020833333333333332  F_A =7.676338940901854E-5
Plan 4      F_C1 =6.662723350575081E-4   F_C2 =0.020833333333333332  F_A =8.262344529228388E-5
Plan 5      F_C1 =4.727751919925607E-4   F_C2 =0.020833333333333332  F_A =5.876351580944249E-5
Plan 6      F_C1 =6.662723350575081E-4   F_C2 =0.020833333333333332  F_A =8.262344529228388E-5
Plan 7      F_C1 =6.662723350575081E-4   F_C2 =0.020833333333333332  F_A =8.262344529228388E-5
Plan 8      F_C1 =4.290402414243473E-4   F_C2 =0.020833333333333332  F_A =5.335533116805599E-5
Plan 9      F_C1 =6.967986411053363E-4   F_C2 =0.02                  F_A =8.6347742765399E-5
```

Figure 13 : L'évaluation de dix plans de la population initiale de l'agent $A_3$.

- La sélection : cette étape consiste à éliminer les moins bons plans de la $Pop_{i0}$ et de garder uniquement les meilleurs en fonction de leur évaluation.
- Le croisement : cette étape consiste à croiser les plans précédemment sélectionnés pour obtenir la nouvelle population $Pop_{pi1}$.

A partir de la population $Pop_{i1}$ obtenue, les étapes décrites plus haut seront réappliquées, dans la limite du nombre de générations possibles précisé par le programme (dans notre cas k=30). Ceci, dans le but d'obtenir une population dans laquelle se trouve un plan qui satisfasse totalement l'ensemble des contraintes. Ce plan sera considéré comme plan initial $p_{i0}$ que l'agent $A_i$ doit exécuter. Dans le cas où ce plan n'est pas obtenu dans la limite des générations possibles, le meilleur plan obtenu dans la dernière génération ($Pop_{ik}$) sera pris en tant que plan initial. Après 30 générations voici les meilleurs plans obtenus (figure 14):

```
P10 = (Move S2,false)(TakeS2,Article 2,150,false) (Move T1,false)
(DelevryT1,Article 2,150,false)(Move S1,false)(TakeS1,Article 1,100,false)
(Move T3,false)(DelevryT3,Article 1,100,false)(Move S3,false)(TakeS3,Article 3,50,false)
(Move T4,false)(DelevryT4,Article 3,50,false)

F_C1 =6.323401357583406E-4   F_C2 =0.01   F_A =7.781241907869489E-5

P20 =(Move S1,false)(TakeS1,Article 1,200,false)(Move S3,false)
(TakeS3,Article 3,150,false)(Move T2,false)(DelevryT2,Article 3,150,false)
(Move S2,false)(TakeS2,Article 2,300,false)(Move T6,false)
(DelevryT6,Article 1,200,false)(Move T7,false)(DelevryT7,Article 2,300,false)
(Move S4,false)(TakeS4,Article 4,50,false)(Move T5,false)(DelevryT5,Article 4,50,false)

F_C1 =3.116105495296216E-4   F_C2 =0.00641025641025641   F_A =3.8483634339865544E-5

P30 = (Move S4,false)(TakeS4,Article 4,200,false) (Move S2,false)(TakeS2,Article
2,150,false)(Move T3,false)(DelevryT3,Article 2,150,false) (Move
S1,false)(TakeS1,Article 1,150,false)(Move T1,false)(DelevryT1,Article 4,200,false)
(Move T2,false)(DelevryT2,Article 1,150,false)

F_C1 =6.662723350575081E-4   F_C2 =0.0208333333333333332   F_A =8.262344529228388E-5
```

Figure 14 : Les meilleurs plans obtenus par les agents $A_1$ $A_2$ $A_3$.

La figure suivante (figure 15) représente les meilleurs plans obtenus par les agents $A_1$ $A_2$ $A_3$ (plan $A_1$ en rouge, plan $A_2$ en bleu, plan $A_3$ en vert)

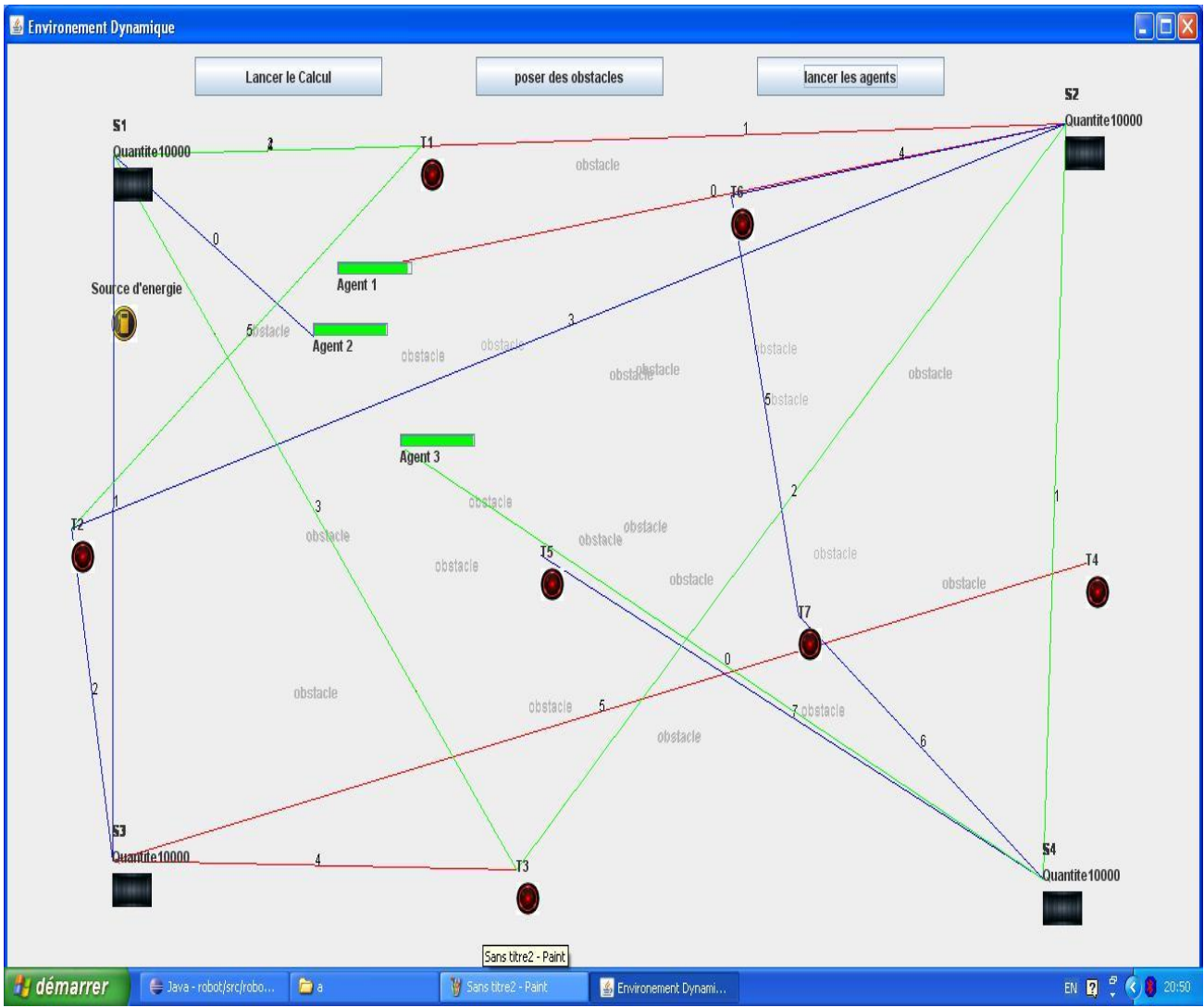

Figure 15 : La représentation des meilleurs plans obtenus par les agents $A_1$ $A_2$ $A_3$.

### 3.2. Génération dynamique des plans après un changement

Supposons que pendant l'exécution des plans initiaux par chaque agent, un nouvel article est introduit dans un nouveau point de stock $S_5$ et que suite à l'introduction de ce nouvel article, les requêtes suivantes ont été formulées:

- $R_{11}$= ($S_5$, $Art_5$, $T_1$), 400, $A_1$, False)
- $R_{12}$=($S_5$, $Art_5$, $T_9$), 100, $A_2$, False)
- $R_{13}$=($S_3$, $Art_5$, $T_8$), 100, $A_3$, False)

Les agents concernés par ce changement doivent générer un nouveau plan dans lequel ils prennent en considération les nouvelles actions nécessaires pour accomplir les nouvelles requêtes. Pour la génération de ce nouveau plan $P_{it}$, l'agent applique les mêmes étapes appliquées pour la génération du plan initial $P_{i0}$ sachant que :

- L'ensemble d'action de chaque agent $\alpha_{it}$ sera égal à l'ensemble des anciennes actions non exécutées de l'ancien plan plus les nouvelles actions engendrées par les changements. les actions réalisées dans le plan précédent sont notées en bleu ; elles ne seront pas prises en compte lors de la génération des nouveaux plans. Les actions non réalisées dans le plan précédent sont notées en noire. Les nouvelles actions engendrées par les changements sont notées en rouge:
    - $\alpha_{1t}$ = {(Move ($S_1$), True), (Take ($S_1$, $Art_1$, 100), True), (Move ($T_3$), True), (Delivery($T_3$, $Art_1$, 100), True), (Move ($S_3$), False), (Take ($S_3$, $Art_3$, 50), False), (Move ($T_4$), False), (Delivery($T_4$, $Art_3$, 50), False), (Move ($S_2$), True), (Take ($S_2$, $Art_2$, 150), True), (Move ($T_1$), True), (Delivery($T_1$, $Art_2$, 150), True), (Move ($S_5$), False), (Take ($S_5$, $Art_5$ 400), False), (Move ($T_1$), False), (Delivery ($T_1$, $Art_5$, 400), False)}
    - $\alpha_{2t}$ = {(Move ($S_3$), True), ( Take ($S_3$, $Art_3$, 150), True), (Move ($T_2$), True), (Delivery($T_2$, $Art_3$, 150), True), (Move ($S_4$), False), (Take ($S_4$, $Art_4$, 50), False), (Move ($T_5$), False), (Delivery($T_5$, $Art_4$, 50), False), (Move ($S_2$), True), (Take ($S_2$, $Art_2$, 300), True), (Move ($T_7$), True), (Delivery($T_7$, $Art_2$, 300), True), (Move ($S_1$), True), (Take ($S_1$, $Art_1$, 200), True), (Move ($T_6$), True), (Delivery($T_6$, Art1, 200), True), Move ($S_5$), False), (Take ($S_5$, $Art_5$, 100), False), (Move ($T_9$), False), (Delivery($T_9$, $Art_5$, 100), False)}
    - $\alpha_{3t}$ = {(Move ($S_2$), True), (Take ($S_2$, $Art_2$, 150), True), (Move ($T_3$), True), (Delivery($T_3$, $Art_2$, 150), True), (Move ($S_1$), False), (Take ($S_1$, $Art_1$, 150), False), (Move ($T_2$), False), (Delivery($T_2$, $Art_1$,150), False), (Move ($S_4$), True), (Take ($S_4$, $Art_4$, 200), True), (Move ($T_1$), False), (Delivery($T_1$, $Art_4$, 200), False), (Move ($S_5$), False), ( Take ($S_5$, $Art_5$, 100), False), (Move ($T_8$), False), (Delivery($T_8$, $Art_5$, 100), False)}
- la population initiale $Pop_{i0}$ de chaque agent $A_i$ sera générée sur la base de son nouvel ensemble d'actions $\alpha_{it}$
- l'état initial de l'agent pour ce plan sera l'état de l'agent à l'instant t $S_{it}$ :
    - $S_{1t}$= {Position (1600,670), Batterie (40%), Taux_Actions_réalisées = 8/16}

- $S_{2t}$ = {Position (490, 570), Batterie (70%), Taux_Actions_réalisées = 12/20}
- $S_{3t}$ = {Position (880, 500), Batterie (80%), Taux_Actions_réalisées = 6 / 16}

- l'état initial de système pour ce plan sera l'état de système à l'instant t $S_{st}$ (les requêtes en bleu représente les requêtes réalisés :

  - $S_{0s}$ = {S= [$S_1$(P=(200, 150), Q=9900), $S_2$(P=(1800, 120), Q=9700), $S_3$(P=(180,1000), Q=9950), $S_4$(P(1750,1010), Q=9800), $S_5$((P=(300, 620), Q=10000)], T= [$T_1$(P=(500,100)), $T_2$(P=(100,800)), $T_3$(P=(800, 800)), $T_4$(P=(1800, 720)), $T_5$(P=(650, 670)) $T_6$(P=(800, 320)) $T_7$(P=(850, 700)) $T_8$(P=(400, 770)) $T_9$(P=(690, 300)) ], R= [$R_1$= (($S_1$, $Art_1$, $T_3$), 100, $A_1$, True), $R_2$=(($S_3$, $Art_3$, $T_4$), 50, $A_1$, False), $R_3$=(($S_2$, $Art_2$, T1), 150, $A_1$, True), $R_4$= (($S_3$, $Art_3$, $T_2$), 150, $A_2$, True), $R_5$=((S4, $Art_4$ ,T5), 50, $A_2$, False), $R_6$=(($S_2$, $Art_2$, $T_7$), 300, $A_2$, True), $R_7$=(($S_1$, $Art_1$, $T_6$), 200, $A_2$, True), $R_8$= (($S_2$, $Art_2$,$T_3$), 150, $A_3$, True), $R_9$=(($S_1$, $Art_1$, $T_2$), 150, $A_3$, False), $R_{10}$=(($S_4$, $Art_4$, $T_1$), 200, $A_3$, False), $R_{11}$= (($S_5$, $Art_5$, T1), 400, $A_1$, False), $R_{12}$=(($S_5$, $Art_5$ , $T_9$), 100, $A_2$, False), $R_{13}$=(($S_3$, $Art_5$, $T_8$), 100, $A_3$, False)]}

L'application de l'approche avec ces nouvelles conditions donne les plans suivants:

```
P1t = (Move S3,false)(TakeS3,Article 3,50,false) (Move S5,false)
(TakeS5,Article 5,400,false) (MoveT1,false)(DelevryT1,Article5,400,false)
(Move T4,false)(DelevryT4,Article 3,50,false

F_C1 =7.323401357583406E-4   F_C2 =0.01   F_A =6.381241907869489E-5

P2t = (Move S4,false)(TakeS4,Article 4,50,false) (Move T5,false)
(DelevryT5,Article4,50,false)(Move S5,false)(TakeS5,Article 5,100,false) (Move T9,false)
(DelevryT9,Article5,100,false)

F_C1 =4.165105495296216E-4   F_C2 =0.00961025641025641   F_A =2.8483634339865544E-5

P3t = (Move T1,false)(DelevryT1,Article 4,200,false) (Move S1,false)
(TakeS1,Article 1,150,false) (Move T2,false)(DelevryT2,Article 1,150,false)
(Move S5,false)(TakeS5,Article 5,100,false) (MoveT8,false)(DelevryT8,Article5,100,false)

F_C1 =8.092723350575081E-4   F_C2 =0.020833333333333332   F_A =5.295344529228388E-5
```

Figure 16 : Les meilleurs nouveaux plans obtenus par les agents $A_1$ $A_2$ $A_3$.

La figure suivante (figure 16) représente les meilleurs nouveaux plans obtenus par les agents $A_1$ $A_2$ $A_3$

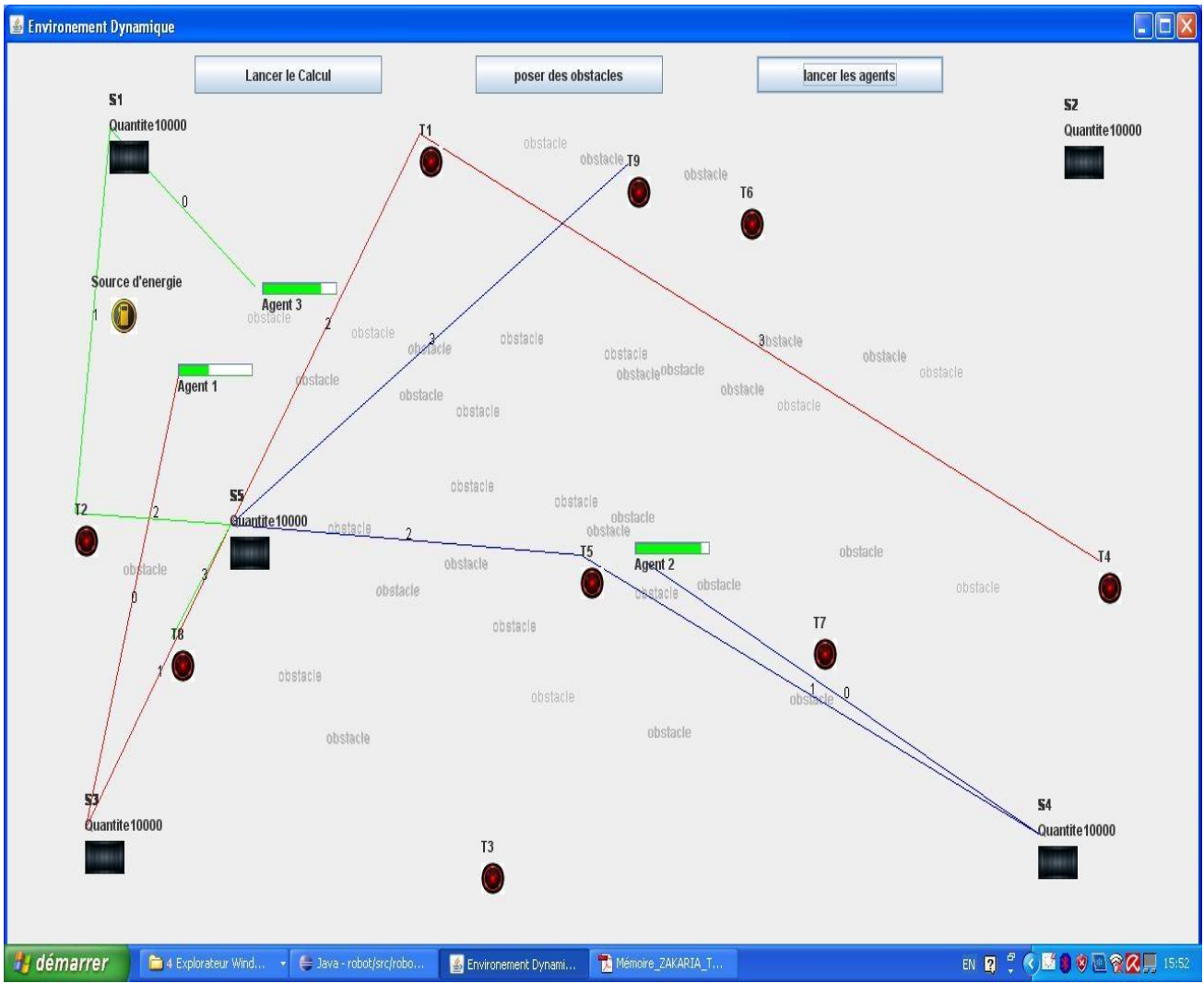

Figure 17 : La représentation des meilleurs nouveaux plans obtenus par les agents $A_1$ $A_2$ $A_3$.

4. **Conclusion**

Dans ce chapitre, nous avons appliqué notre approche de planification sur un cas concret, ce qui nous a permis de montrer que notre approche offre plusieurs avantages, à savoir:

- La prise en compte des changements pouvant survenir sur l'ensemble des actions à planifier par chaque agent suite à la prise en compte des changements observables dans son environnement

- Evite de revenir au point zéro pour la génération d'un nouveau plan, puisque l'agent prend en considération uniquement les anciennes actions non exécutées de l'ancien plan auxquelles il ajoute les nouvelles actions engendrées par les changements.
- Assure que le plan généré, à un instant donné, est le meilleur plan grâce à l'utilisation des algorithmes génétique.
- L'approche est indépendante de l'architecture de l'agent.
- Chaque agent génère ses propres plans indépendamment des autres agents ce qui minimise le coût de communications et assure une grande flexibilité ; en effet si un agent tombe en panne le système continue à fonctionner.

# Conclusion générale

La technologie orientée-agent est devenue un paradigme à part entière du génie logiciel disposant de ses propres éléments méthodologiques en termes de conception et de programmation. Le succès de ce paradigme, par rapport aux autres paradigmes, repose sur ses caractéristiques fonctionnelles et comportementales propres telles que l'autonomie, la proactivité, la sensibilité, la flexibilité, etc.

L'extension de la planification dans le cadre des systèmes multi-agent a aboutit à la planification distribuée. Cette extension a enrichi le domaine de la planification suite à l'exploitation des avantages du paradigme multi-agent. Cependant, l'utilisation de ce dernier a introduit de nouveaux défis inexistants dans la version classique de la planification. En effet dans la planification classique, l'ensemble des actions à planifier est défini auparavant et ne subit aucun changement assurant ainsi, une fiabilité du plan généré jusqu'à la fin de son exécution. Alors que dans la planification distribuée, chaque agent peut faire des changements dans son ensemble d'actions à planifier, afin de prendre en considération les changements imprédictible de son environnement.

Dans ce mémoire, nous avons proposé une nouvelle approche de planification distribuée dynamique, capable de prendre en considération les changements que l'agent introduit sur son ensemble des actions à planifier dans le but de prendre en compte les changements qui surviennent dans son environnement

Selon notre approche chaque agent commence par la génération de son plan initial, dans lequel il établit le meilleur ordre, de l'ensemble des actions initiales qu'il doit exécuter afin de répondre de la manière la plus satisfaisante à l'ensemble de contraintes. Pendant l'exécution du plan généré, si l'agent se retrouve devant un changement, vis-à-vis duquel, de nouvelles actions doivent être exécutées, un autre plan, sera généré d'une façon dynamique pour prendre en considération ces actions. Dans ce nouveau plan, l'agent doit établir le meilleur ordre entre les anciennes actions non exécutées de l'ancien plan et les nouvelles actions engendrées par les changements de façon à satisfaire l'ensemble des contraintes. De ce fait, la génération des plans se répète d'une façon récursive, prenant, à chaque fois, comme nouvel état initial; l'état dans lequel l'ensemble des actions de l'agent subit un changement.

L'approche proposée offre plusieurs avantages, à savoir:

- La prise en compte des changements pouvant survenir sur l'ensemble des actions à planifier par chaque agent suite à la prise en compte des changements observables dans son environnement.
- Evite de revenir au point zéro pour la génération d'un nouveau plan, puisque l'agent prend en considération uniquement les anciennes actions non exécutées de l'ancien plan auxquelles il ajoute les nouvelles actions engendrées par les changements.
- Assure que le plan généré, à un instant donné, est le meilleur plan grâce à l'utilisation des algorithmes génétique.
- L'approche est indépendante de l'architecture de l'agent.
- Chaque agent génère ses propres plans indépendamment des autres agents ce qui minimise le coût de communications et assure une grande flexibilité ; en effet si un agent tombe en panne le système continue à fonctionner.

L'approche proposée, a été validée sur une étude de cas concret: DPDP. Selon les résultats obtenus, il serait intéressant d'intégrer notre approche dans les plateformes de développement des agents comme bibliothèque à part pour supporter le processus de planification.

Comme perspectives, nous prévoyons à court et moyen termes l'optimisation du temps de calcul de génération des plans ainsi que la prise en compte des changements des contraintes.

# Bibliographie